\documentclass{article} 
\usepackage{iclr2027_conference,times}

\usepackage{amsmath,amsfonts,bm}

\def\eqref#1{equation~\ref{#1}}

\def\1{\bm{1}}

\DeclareMathAlphabet{\mathsfit}{\encodingdefault}{\sfdefault}{m}{sl}
\SetMathAlphabet{\mathsfit}{bold}{\encodingdefault}{\sfdefault}{bx}{n}

\usepackage{hyperref}
\usepackage{url}

\usepackage{graphicx}
\usepackage{booktabs}
\usepackage{amsmath}
\usepackage{amsfonts}
\usepackage{array}
\usepackage{multirow}
\usepackage{adjustbox}
\usepackage{algorithm}
\usepackage{algpseudocode}
\usepackage{tabularx}
\usepackage[table]{xcolor}
\definecolor{AoLBlue}{RGB}{241,246,255}
\definecolor{HeaderGray}{RGB}{247,247,247}

\title{Age of Learning: Temporal Persistence of \\Prediction Errors as a Learning Signal}

\author{Chenyang Wang, Stefan Forsström \& Roger Olsson \\
Mid Sweden University \\
\texttt{\{chenyang.wang,stefan.forsstrom,roger.olsson\}@miun.se} \\
\AND
Di Yuan \\
Uppsala University \\
\texttt{di.yuan@it.uu.se} \\
\And
Qing He \\
Mid Sweden University \\
\texttt{qing.he@miun.se}
}

\iclrfinalcopy 
\begin{document}

\maketitle

\begingroup
\renewcommand{\thefootnote}{\phantom{0}}
\footnotetext[1]{Code: \url{https://github.com/sanyeungwang/Age-of-Learning}}
\endgroup

\lhead{Preprint}

\begin{abstract}
Current machine learning algorithms primarily rely on instantaneous signals such as loss, margin, and prediction confidence to characterize model behavior. These signals indicate how difficult a prediction is at the current optimization step, but they do not capture how long the model has remained incorrect. We study this temporal dimension of learning and introduce Age of Learning (AoL), a learning-state variable that measures the persistence of prediction errors over time. AoL increases while an error remains unresolved and resets when a correct prediction is achieved, thereby distinguishing persistent under-learning from transient mistakes. We develop AoL-based training strategies for both offline and streaming settings. In offline learning, sample-level AoL is accumulated over training and aggregated into class-level states that guide adaptive reweighting and resampling. In streaming learning, where full historical access is unavailable, we maintain lightweight class-level AoL states using current and buffered observations. Across long-tailed classification settings, AoL improves or matches standard training baselines, with larger benefits when learning difficulty persists over time. Multi-seed streaming experiments further show reproducible gains under temporally stable imbalance. Analysis of class frequency, loss, and margin shows that AoL is related to conventional difficulty measures but captures additional information about error duration. These results suggest that temporal persistence provides a useful complementary signal for characterizing and controlling learning dynamics in imbalanced and non-stationary environments.
\end{abstract}

\section{Introduction}
\label{sec:introduction}

Modern machine learning and deep learning models~\citep{bishop2006pattern, Goodfellow-et-al-2016} are commonly trained using instantaneous signals such as loss~\citep{Cui_2019_CVPR}, margin~\citep{NEURIPS2019_621461af}, and prediction confidence~\citep{Lin_2017_ICCV}, which quantify how well a model currently fits a sample, while dataset-level statistics such as class frequency characterize the training distribution. These signals support a wide range of learning strategies, including class reweighting~\citep{Cui_2019_CVPR}, hard-example mining~\citep{Lin_2017_ICCV}, margin-based objectives, and adaptive sampling~\citep{pmlr-v267-hacohen25a}. However, they leave out a different aspect of learning: how long a prediction error has remained unresolved.

This temporal information can matter when learning difficulty persists across optimization steps. Consider two samples that are currently misclassified with similar loss or margin. One may have only recently become incorrect, whereas the other may have remained misclassified for many consecutive epochs. Instantaneous difficulty measures treat these cases similarly despite their different learning histories. The same issue arises at the class level: classes with similar frequencies or current losses may exhibit very different histories of persistent under-learning. Such behavior is particularly relevant in long-tailed learning~\citep{Cui_2019_CVPR, Liu_2019_CVPR, NEURIPS2019_621461af} and streaming data~\citep{NEURIPS2019_e562cd9c, NEURIPS2024_d8f5f134, wang2026streaming}. In long-tailed datasets~\citep{10105457}, rare classes may remain misclassified for extended periods due to insufficient representation, while in streaming scenarios certain classes may be neglected or become stale as the data distribution evolves. In both cases, it is not only the error magnitude that matters, but also its temporal persistence across successive optimization steps throughout training.

\begin{figure}[htbp]
  \centering
  \begin{minipage}[c]{0.7\linewidth}
    \centering
    \includegraphics[width=\linewidth]{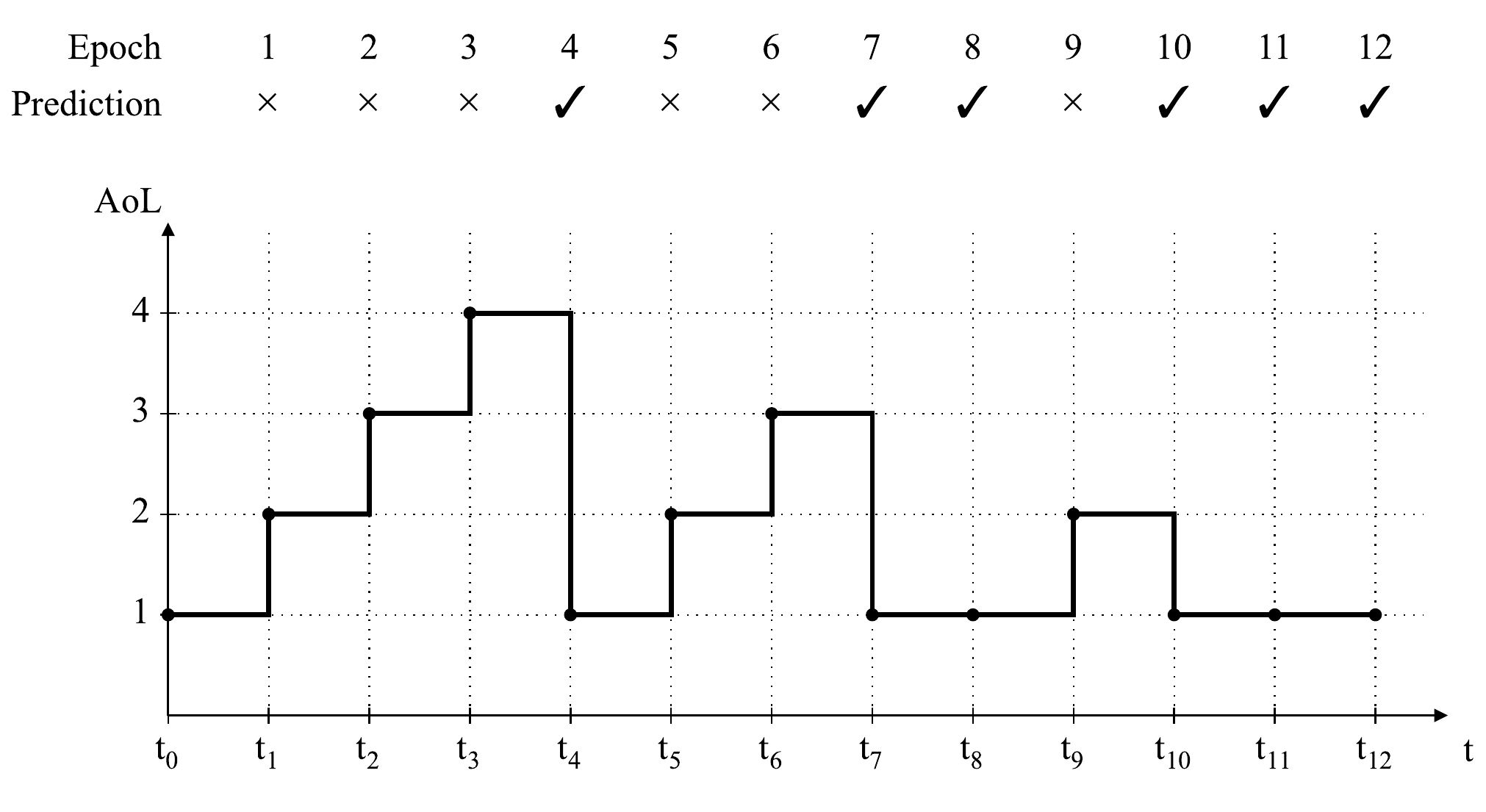}
  \end{minipage}\hfill
  \begin{minipage}[c]{0.28\linewidth}
    \caption{Illustration of Age of Learning (AoL). AoL increases when a sample or class remains misclassified and resets when correct classification is achieved. Rather than measuring instantaneous error severity, AoL captures error duration as a temporal learning-state variable that evolves with model updates to guide subsequent optimization.}
    \label{fig:AoL}
  \end{minipage}
\end{figure}

This work studies error persistence as an explicit learning signal. We introduce Age of Learning (AoL), a temporal learning-state variable that measures how long prediction error persists during training. For each sample or class, AoL increases when misclassification occurs and resets when correct classification is achieved. We illustrate the core idea in Figure~\ref{fig:AoL}.  

AoL complements instantaneous difficulty signals by measuring how long prediction errors persist. A closely related temporal concept is Age of Incorrect Information (AoII)~\citep{9137714}, which measures how long a receiver’s estimate remains incorrect relative to an external source in communication systems. In contrast, AoL defines an endogenous temporal state of the learning process itself: it is derived from a trainable model’s prediction trajectory, evolves with model updates, and feeds back into subsequent optimization. This enables persistent prediction error to serve directly as a learning-state variable for adaptive training.

We first formulate AoL in an offline setting, where sample-level AoL is computed and aggregated into class-level signals that guide training, particularly in long-tail scenarios. We then extend AoL to streaming learning, where full historical access is unavailable and sample-level tracking is infeasible, by maintaining lightweight per-class states updated from the current stream and optional buffered data. We evaluate AoL on long-tailed classification tasks in both offline and streaming settings. The results show that, in offline settings, AoL-based training improves or matches standard baselines; in streaming scenarios, it enables adaptive reweighting based on persistent misclassification and shows clear advantages when learning difficulty evolves with temporal consistency. More broadly, our results suggest that temporal behavior can serve as a useful optimization signal beyond static dataset statistics or instantaneous prediction errors. The main contributions of this work are:

\begin{itemize}
    \item We introduce AoL, a temporal learning-state variable that explicitly represents the persistence of prediction errors.
    \item We develop AoL formulations for both offline and streaming learning, including sample-level and class-level states and lightweight online updates under restricted historical access.
    \item We show how AoL can be incorporated into adaptive training through loss reweighting and weighted resampling, providing concrete mechanisms for testing whether temporal persistence is useful for optimization.
    \item We empirically analyze the relationship between AoL and class frequency, loss, and margin, and show that AoL captures temporal information not determined by these conventional signals.
    \item We demonstrate reproducible improvements in long-tailed offline and streaming settings, with the strongest benefits arising when learning difficulty itself persists over time.
\end{itemize}

\section{Related Work}
\label{sec:related_work}

We review prior work most relevant to AoL, including long-tail learning, sample reweighting, online learning, learning dynamics, and age-based metrics, and clarify the positioning of AoL relative to these areas across offline and streaming settings.

\textbf{Long-tail learning.}
Long-tail learning addresses highly imbalanced class distributions~\citep{Zhang_2021_ICCV, Alshammari_2022_CVPR, NEURIPS2023_eeffa70b, NEURIPS2024_350e718f, 10.1007/978-3-031-72949-2_18, pmlr-v267-li25dk, cortes2026improved}. Existing approaches include class-sensitive loss design and reweighting based on label frequency or effective sample size, margin-based objectives~\citep{Cui_2019_CVPR, NEURIPS2019_621461af}, decoupled representation and classifier learning~\citep{Kang2020Decoupling}, and re-sampling or hard-example emphasis~\citep{Shrivastava_2016_CVPR, Lin_2017_ICCV, 10.1007/978-3-030-65414-6_9, NEURIPS2023_eeffa70b}. These approaches primarily address imbalance through class statistics, margin design, or data-level interventions. AoL addresses a different dimension: classes with similar frequencies may exhibit very different learning trajectories, and AoL measures how long under-learning persists during optimization rather than inferring importance from sample count alone.

\textbf{Sample reweighting.}
Example-weighting and selection methods construct importance from various signals. Meta-reweighting methods learn weights through meta-gradients and/or a clean validation or meta set~\citep{pmlr-v80-ren18a,NEURIPS2019_e58cc5ca}; optimal-transport reweighting targets distributional rebalancing towards a balanced meta distribution~\citep{NEURIPS2022_a39a9ace}; importance sampling primarily reduces stochastic-gradient variance while maintaining an unbiased estimator of the original objective~\citep{pmlr-v80-katharopoulos18a}. Other approaches construct weighting from loss-based objectives or related instantaneous difficulty signals~\citep{yi2021reweighting,ICLR2025_ded26b34}, while data-selection methods use early-training gradient or error scores~\citep{NEURIPS2021_ac56f8fe}. AoL instead derives importance from the duration of unresolved prediction errors across updates rather than from static statistics or instantaneous difficulty. We incorporate this temporal signal into training through loss reweighting and weighted resampling.

\textbf{Online learning.}
Online and streaming learning studies how models can be updated when data arrive sequentially, often under strict memory and access constraints~\citep{10.1007/978-3-030-58604-1_43, pmlr-v139-flaspohler21a, wu2024stabilizing, zhang2025nonstationary, sridharan2025online, daniely2026online, liu2026online}. A large body of work addresses catastrophic forgetting~\citep{NIPS2017_f8752278, doi:10.1073/pnas.1611835114} and non-stationarity~\citep{NIPS2003_feecee9f, zhang2025nonstationary} through replay mechanisms~\citep{NEURIPS2019_e562cd9c}, memory buffers~\citep{pmlr-v267-hacohen25a}, or continual-learning regularization~\citep{NEURIPS2024_d8f5f134}. Replay-based methods retain past samples to support learning. Our streaming formulation maintains a class-level AoL state to guide training emphasis and can use a replay buffer to estimate current class performance. In long-tailed streams, infrequently observed classes may remain under-learned for long periods.

\textbf{Learning dynamics.}
Prior work studies temporal learning behavior through forgetting events and related sample-level dynamics~\citep{toneva2018an,pmlr-v267-hacohen25a,zheng2026understanding,jin2026orderdp}. Example-forgetting methods record transitions between learned and forgotten states and show that prediction trajectories reveal information beyond final loss or accuracy. AoL shares this view but measures a different property: forgetting statistics characterize transitions between correctness states, whereas AoL measures the duration of consecutive incorrectness. Repeated short error episodes and one long uninterrupted error episode may therefore produce similar transition counts but substantially different AoL trajectories.

\textbf{Other age-based metrics.}
Temporal notions of ``age'' have also been studied in communication and status-update systems~\citep{8000687,8514816,8734015,9380899}. Age of Information (AoI)~\citep{6195689,7364263} measures the freshness of received information, while Age of Incorrect Information (AoII)~\citep{9137714} measures how long a receiver estimate remains incorrect relative to an external process. AoII is the closest conceptual antecedent to AoL because both assign temporal significance to persistent incorrectness. However, the state, dynamics, and role are different: AoII is defined by disagreement between an external source and a receiver estimate and is used to guide communication or update policies, whereas AoL is an endogenous learning-state variable derived from a trainable model's prediction trajectory relative to supervision and feeds back into subsequent optimization. AoL formulates persistent prediction error as a temporal state for characterizing and guiding learning.

\section{Age of Learning}
\label{sec:approach}

In this section, we formalize AoL as a temporal learning-state variable and describe how it can be incorporated into model training. We first introduce the quantities in the standard offline setting and then extend the formulation to streaming learning.

\subsection{Offline AoL}

We first define AoL in an offline setting, where the full dataset is accessible during training. Let $i\in\{1,\dots,N\}$ index training samples, and $e\in\{1,\dots,E\}$ denote the training epoch index. The model prediction for sample $i$ at epoch $e$ is given by:
\begin{equation}
    \hat{y}_i^{(e)}=\arg\max_k f_{\theta^{(e)}}(\boldsymbol{x}_i)_k, \quad k\in\{1,\dots,C\}
\end{equation}
where $f_{\theta^{(e)}} : \mathcal{X} \rightarrow \mathbb{R}^C$ denotes a neural network parameterized by $\theta^{(e)}$ at epoch \(e\), which maps an input $\boldsymbol{x}_i$ to a $C$-dimensional logit vector, where $C$ is the total number of classes, and $f_{\theta^{(e)}}(\boldsymbol{x}_i)_k$ represents the logit corresponding to class $k$. The predicted label $\hat{y}_i^{(e)}$ is therefore the class index with the maximum logit value. Next, we define a binary correctness indicator
$c_i^{(e)} = \mathbb{I}[\hat{y}_i^{(e)} = y_i]$, where $y_i \in \{1, \dots, C\}$ denotes the ground-truth label of sample $i$ and $\mathbb{I}[\cdot]$ denotes the indicator function. Thus, $c_i^{(e)} = 1$ indicates that the model correctly classifies sample $i$ at epoch $e$, while $c_i^{(e)} = 0$ indicates a misclassification.

At the beginning of training, the AoL of sample $i$ is initialized as one. At the end of each epoch, the AoL is updated based on the prediction correctness such that $a_i^{(e)} = 1$ if the sample is correctly classified, otherwise $a_i^{(e)} = a_i^{(e-1)} + 1$, as follows:
\begin{equation}
a_i^{(e)} =
\begin{cases}
1, & \text{if } c_i^{(e)} = 1, \\
a_i^{(e-1)} + 1, & \text{otherwise},
\end{cases}
\end{equation}
which can be equivalently written in a compact form as:
\begin{equation}
a_i^{(e)} = (1 - c_i^{(e)})(a_i^{(e-1)} + 1) + c_i^{(e)}.
\end{equation}

By definition, the instantaneous AoL state $a_i^{(e)}$ records the duration of the current consecutive misclassification episode. Here we use a one-based convention in which $1$ is the minimum AoL state. This gives all samples equal positive initial states and avoids a degenerate zero denominator when AoL is subsequently normalized to construct training weights.

Instantaneous AoL may fluctuate as predictions switch between correct and incorrect states. To obtain a smoother learning-state variable that retains information from previous error episodes, we maintain an aggregated AoL, denoted by $\bar{a}_i^{(e)}$, using an exponential moving average (EMA):
\begin{align}
\bar{a}_i^{(1)}
&=
a_i^{(1)},
\label{eq:ema_init}
\\
\bar{a}_i^{(e)}
&=
(1-\rho)\bar{a}_i^{(e-1)}
+
\rho a_i^{(e)},
\qquad e\geq 2,
\label{eq:ema_aol}
\end{align}
where $\rho\in[0,1]$ controls the trade-off between historical memory and responsiveness to the current error episode. A larger $\rho$ makes $\bar{a}_i^{(e)}$ respond more strongly to recent prediction behavior, whereas a smaller $\rho$ produces a smoother state with longer temporal memory. The two limiting cases are illustrative: when $\rho=1$, then $\bar{a}_i^{(e)} = a_i^{(e)}$, only the current error episode is retained; when $\rho=0$, then $\bar{a}_i^{(e)} = \bar{a}_i^{(1)}$, and the aggregated state remains fixed at its initialization. Thus, nonzero intermediate values of $\rho$ explicitly combine current error persistence with historical learning behavior.

Compared to the instantaneous AoL $a_i^{(e)}$, which measures \emph{how long the current incorrect state has persisted}, the aggregated AoL $\bar{a}_i^{(e)}$ summarizes \emph{how persistent incorrectness has been over the learning process}. In the following section, we use the aggregated AoL to construct sample-level and class-level training signals.

\subsection{AoL-Guided Reweighting and Resampling}
We now describe how the resulting AoL signals are used to guide the optimization process. Since larger AoL values indicate persistent misclassifications, the training process should prioritize samples or classes with higher AoL to improve their contribution to the model. The aggregated AoL $\bar{a}_i^{(e)}$ serves as an indicator for the accumulated misclassifications, and it is subsequently used to construct training weights. To ensure scale invariance, we normalize the aggregated AoL across all $N$ training samples and define a sample-level weight for each training sample $i$ as:
\begin{equation}
    w_i^{(e)} = \frac{\bar{a}_i^{(e)}}{\frac{1}{N}\sum_{j=1}^{N} \bar{a}_j^{(e)}}.
\end{equation}
This normalization preserves the overall magnitude of the loss while relatively emphasizing samples with larger AoL values, i.e., those whose misclassifications have remained for longer periods.

Furthermore, we consider a class-level formulation, which is the primary setting used in our method. Let $\mathcal{I}_c$ denote the set of samples belonging to class $c$, and $|\mathcal{I}_c|$ denote the number of elements in $\mathcal{I}_c$. The class-level AoL is defined as:
\begin{equation}
    \bar{a}_c^{(e)} = \frac{1}{|\mathcal{I}_c|} \sum_{i \in \mathcal{I}_c} \bar{a}_i^{(e)},
\end{equation}
which represents the average aggregated AoL over all samples in class $c$. Meanwhile, the corresponding class-level weight for each class $c$ is defined as:
\begin{equation}
    w_c^{(e)} = \frac{\bar{a}_c^{(e)}}{\frac{1}{C}\sum_{k=1}^{C} \bar{a}_k^{(e)}}.
\label{eq:class_weight}
\end{equation}

Compared to the sample-level weights, which directly reflect individual sample dynamics but may suffer from high variance, the class-level weights aggregate information across samples within the same class and therefore provide a more stable signal in long-tailed settings. In practice, both variants are evaluated in the experiments, and the latter generally yields more consistent improvements.

Next we incorporate the AoL-based weights into the training objective. We consider two commonly used classification losses: Softmax cross-entropy and Sigmoid cross-entropy. For the Softmax cross-entropy, the per-sample loss is defined as:
\begin{equation}
\ell_i^{\text{Softmax}} = -\log \frac{\exp(z_{i,y_i})}{\sum_{k=1}^{C} \exp(z_{i,k})},
\end{equation}
where $z_{i,k} = f_{\theta}(\boldsymbol{x}_i)_k$ denotes the logit corresponding to class $k$. For the Sigmoid cross-entropy, we adopt a one-vs-all formulation:
\begin{equation}
\ell_i^{\text{Sigmoid}} = - \sum_{k=1}^{C} \Big[ y_{i,k} \log \sigma(z_{i,k}) + (1 - y_{i,k}) \log (1 - \sigma(z_{i,k})) \Big],
\end{equation}
where $y_{i,k}$ is the one-hot label and $\sigma(\cdot)$ is the Sigmoid function. 

To incorporate the AoL information, we apply loss reweighting by multiplying each sample's loss with its corresponding weight. This can be written in a unified form as:
\begin{equation}
\mathcal{L}^{(e)} = \frac{\sum_{i \in \mathcal{B}} w_i^{(e)} \, \ell_i}{\sum_{i \in \mathcal{B}} w_i^{(e)}},
\end{equation}
where $\ell_i$ denotes either $\ell_i^{\text{Softmax}}$ or $\ell_i^{\text{Sigmoid}}$, and $\mathcal{B}$ denotes the mini-batch. When using class-level weighting, the weight is given by $w_i^{(e)} = w_{y_i}^{(e)}$, where $i \in \{1,\dots,N\}, y_i \in \{1,\dots,C\}$. This assigns the same weight to all samples belonging to one class, i.e., $w_i^{(e)} = w_c^{(e)}$ for all $i \in \mathcal{I}_c$.

In addition to loss reweighting, we also consider incorporating AoL-based weighting into the data sampling process. Specifically, we employ a weighted random sampler with replacement, where the sampling probability of each sample $i$ at epoch $e$ is proportional to its weight $w_i^{(e)}$. In the sample-level setting, $w_i^{(e)}$ is directly used as the sampling weight. In the class-level setting, the sampling weight is obtained by assigning the corresponding class weight to each sample, i.e., $w_i^{(e)} = w_{y_i}^{(e)}$.

Overall, loss reweighting adjusts the contribution of each sample in the objective function, while weighted resampling modifies the data distribution seen by the model. Both mechanisms are guided by the AoL-based signals, encouraging the model to prioritize samples or classes whose incorrectness has become persistent. Computing offline AoL requires one additional forward-only evaluation per epoch, together with $O(N)$ sample-level states and $O(C)$ class-level states. The current and aggregated AoL can be maintained incrementally, without storing the full prediction history or requiring an additional backward pass.

\subsection{Online AoL}
\label{sec:online_aol}

We now extend AoL to streaming learning, where data arrive sequentially and the full training set is no longer repeatedly accessible. In this setting, maintaining a persistent sample-level state for every previously observed example is generally impractical. To address this, we introduce an online formulation of AoL that maintains a lightweight class-level state using the current data stream together with an optional memory buffer.

We use $t\in\{1,\ldots,T\}$ to denote the streaming update step, in contrast to epoch index $e$ used in the offline training. Let $\mathcal{B}^{(t)}$ denote the mini-batch observed at step $t$, and $\mathcal{M}^{(t)}$ a memory buffer containing a subset of previously observed samples. The buffer is updated according to a chosen memory policy and is subject to a fixed storage budget. We define the observation set used to estimate the current class-level learning status as:
\begin{equation}
\mathcal{S}^{(t)}
=
\begin{cases}
\mathcal{M}^{(t)}, & \text{if a memory buffer is used},\\
\mathcal{B}^{(t)}, & \text{otherwise}.
\end{cases}
\label{eq:online_observation_set}
\end{equation}
For each class $c$ represented in $\mathcal{S}^{(t)}$, we compute the current class-level classification accuracy as:
\begin{equation}
r_c^{(t)}
=
\frac{
\sum_{(x,y)\in\mathcal{S}^{(t)}}
\mathbb{I}[y=c]\mathbb{I}[\hat{y}^{(t)}(x)=c]
}{
\sum_{(x,y)\in\mathcal{S}^{(t)}}
\mathbb{I}[y=c]
},
\label{eq:online_class_accuracy}
\end{equation}
where $\hat{y}^{(t)}(x)$ denotes the prediction of the current model at step $t$ for input $x$.

Unlike the offline formulation, where sample-level error episodes can be tracked explicitly, the streaming setting provides only partial and time-varying observations of each class. We therefore update a class-level AoL state according to whether the estimated class accuracy falls below a threshold $\tau^{(t)}$:
\begin{equation}
a_c^{(t)}
=
\begin{cases}
a_c^{(t-1)} + 1,
& \text{if } r_c^{(t)} < \tau^{(t)}, \\[2mm]
1,
& \text{if } r_c^{(t)} \geq \tau^{(t)}, \\[2mm]
a_c^{(t-1)},
& \text{if class } c \text{ is not observed in } \mathcal{S}^{(t)}.
\end{cases}
\label{eq:online_aol}
\end{equation}

All class-level states are initialized as one. The threshold $\tau^{(t)}$ specifies the accuracy level at which an under-learning episode is considered to have ended. It may be chosen according to task requirements or adapted to the observed data. If an observed class has estimated accuracy below $\tau^{(t)}$, its class-level AoL increases, indicating persistent under-learning. Once its estimated accuracy reaches or exceeds $\tau^{(t)}$, the current under-learning episode is considered resolved, and the state resets to one. If the class is not observed, its state remains unchanged. Thus, persistence is measured over observed class evaluations, rather than elapsed streaming time.

To smooth fluctuations in the estimated class-level learning status, we obtain online class weights by applying the same EMA aggregation and mean normalization as in the offline setting to the class-level AoL states. Each observed sample is then weighted by its ground-truth class for online loss reweighting. This online AoL formulation provides a simple and practical mechanism for updating class-level learning states in streaming scenarios. It requires only $O(C)$ additional AoL state beyond the memory buffer and can operate directly on the current mini-batch when no buffer is available. Its computational overhead depends on the size of the observation set $\mathcal{S}^{(t)}$ and the frequency of AoL evaluations. More algorithmic details are provided in Algorithm~\ref{alg:online_aol}, Appendix~\ref{sec:algorithm}.

\section{Experiments}
\label{sec:experiment}

In this section, we first examine the information captured by AoL, then evaluate it in offline and streaming long-tailed settings, including both stable and dynamically evolving streams.

\subsection{Motivating Observation}
\label{sec:exp_observation}

We first examine whether AoL captures information beyond class frequency and conventional instantaneous difficulty signals. We use CIFAR-100~\citep{krizhevsky2009learning} with imbalance factor (IF) $1$ and $200$, representing balanced and highly imbalanced settings, respectively. The long-tailed training set is constructed as:
\begin{equation}
\label{eq:lt}
    n_r = n_{\max}\mu^{-\frac{r}{(C-1)}}, \quad r = 0,\dots,C-1,
\end{equation}
where \(n_r\) is the number of training samples for the class at long-tail rank \(r\), \(n_{\max}\) is the number of samples in the head class, \(C\) is the number of classes, and \(\mu\) is the target IF. We train ResNet-34~\citep{He_2016_CVPR} for 100 epochs with batch size 128.
\begin{figure}[tbp]
  \centering
  \begin{minipage}[c]{0.62\linewidth}
    \centering
    \includegraphics[width=\linewidth]{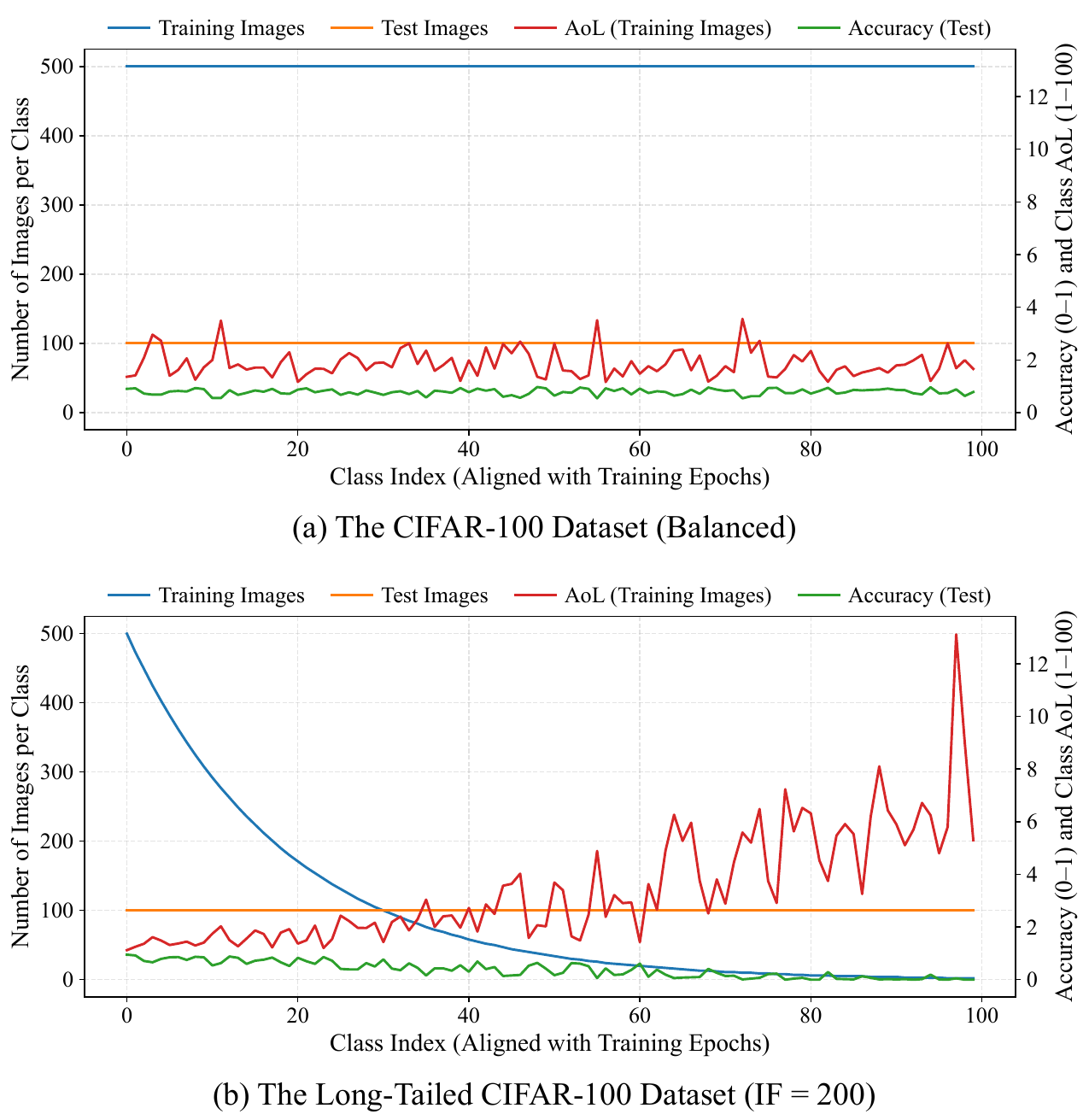}
  \end{minipage}\hfill
  \begin{minipage}[c]{0.35\linewidth}
  \caption{Relationship between class frequency, class-level AoL, and per-class test accuracy on CIFAR-100 under (a) balanced IF$=$1 and (b) long-tailed IF$=$200 settings. The blue and orange curves represent the numbers of training and test images per class (left \(y\)-axis). The red curve shows class-level AoL within \([1,100]\), and the green curve shows the final per-class test accuracy within \([0,1]\) (right \(y\)-axis). Under IF$=$1, all classes have identical training frequency but exhibit different AoL values. Under IF$=$200, it is highly apparent that accuracy is negatively related to AoL, and classes with similar tail frequencies also exhibit different AoL states, indicating that AoL captures variation in persistent under-learning beyond class frequency alone.}
  \label{fig:insight}
  \end{minipage}
\end{figure}

Figure~\ref{fig:insight} compares class frequency, class-level AoL, and final per-class test accuracy. Under IF$=$1, all classes have identical training frequencies but different AoL values, showing that class frequency alone does not determine the temporal learning state. This pattern becomes more pronounced under IF$=$200, where several tail classes with identical or nearly identical sample counts exhibit markedly different AoL values. Thus, prediction-error persistence can differ even among classes with comparable data scarcity. Across both settings, larger class-level AoL tends to correspond to lower final test accuracy, with Pearson correlations of -0.8231 for IF$=$1 and -0.8225 for IF$=$200. These strong negative correlations show that AoL captures persistent under-learning beyond class counts alone.

\paragraph{Correlation analysis.}
We further compare AoL with conventional instantaneous difficulty signals. For IF-$1$, the Pearson correlations of class-level AoL with class-level training loss/margin are 0.7841/-0.7757, and those with class-level test loss/margin are 0.6352/-0.6882. For IF-$200$, the corresponding correlations are 0.6697/-0.6633 on the training set and 0.4300/-0.3883 on the test set. Persistently difficult classes therefore tend to have larger losses and smaller margins, but the correlations are not perfect: loss and margin characterize the current prediction state, whereas AoL additionally captures how long incorrectness persists. Hence, AoL is related to, but distinct from, class frequency, loss, and margin. More detailed AoL dynamics are provided in Appendix~\ref{sec:visualization}.

\subsection{Offline Settings}
\label{sec:exp_offline}

We next evaluate AoL in offline long-tailed classification against standard Softmax and Sigmoid losses and their class-balanced (CB) counterparts. For fair and reproducible comparison, we conduct a PyTorch~\citep{10.5555/3454287.3455008} reimplementation based on the official code of CB~\citep{Cui_2019_CVPR}. We train CIFAR ResNet-32 for \(200\) epochs using SGD with momentum \(0.9\), weight decay \(2\times10^{-4}\), batch size \(128\), and an initial learning rate of \(0.1\), with \(5\)-epoch warmup and step decay at epochs \(160\) and \(180\). All CIFAR-based experiments run on an NVIDIA TITAN X (Pascal) GPU. Training uses random crop and horizontal flip, while AoL is computed on the corresponding unaugmented training-evaluation set to avoid augmentation noise in its updates. Following the long-tailed CIFAR protocol, each imbalanced training set is constructed using the exponential rule in Section~\ref{sec:exp_observation}. To isolate temporal persistence from CB weighting, the main experiments evaluate AoL independently of CB and focus on loss reweighting, where AoL adjusts each training example's contribution to the objective. Table~\ref{tab:offline_main} reports the selected AoL results, while additional results for both sample-level and class-level AoL are provided in Appendix~\ref{app:offline_online_settings}.

\begin{table}[tbp]
\centering
\caption{Offline long-tail classification results on CIFAR-100 and CIFAR-10 under different IFs. Top-1 test accuracy (\%) is reported. The better result between the sample- and class-level AoL is presented in the table. Superscript $\mathrm{s}$ denotes that the result is achieved at the sample-level; the results without the superscript are obtained at the class-level.}
\label{tab:offline_main}

\scriptsize
\renewcommand{\arraystretch}{1.10}
\setlength{\tabcolsep}{3pt}

\begin{minipage}[t]{0.49\linewidth}
\centering
\textbf{CIFAR-100}

\begin{tabularx}{\linewidth}{
l
*{4}{>{\centering\arraybackslash}X}
}
\toprule

\rowcolor{HeaderGray}
\textbf{Method}
& \textbf{IF-400}
& \textbf{IF-100}
& \textbf{IF-10}
& \textbf{IF-1} \\

\midrule

Softmax
& 31.50
& 40.10
& 57.15
& 71.59 \\

Softmax + CB
& 32.04
& 40.40
& 57.48
& 71.59 \\

\rowcolor{AoLBlue}
\textbf{Softmax + AoL}
& \textbf{32.43}
& \textbf{40.51}
& \textbf{57.98}
& \textbf{71.67} \\

\midrule

Sigmoid
& \textbf{33.00}
& 40.65
& 57.63
& 71.33 \\

Sigmoid + CB
& 32.86
& 40.81
& \textbf{58.44}
& 71.33 \\

\rowcolor{AoLBlue}
\textbf{Sigmoid + AoL}
& \textbf{33.00}
& \textbf{40.84}
& 58.07
& \textbf{71.56} \\

\bottomrule
\end{tabularx}
\end{minipage}
\hfill
\begin{minipage}[t]{0.49\linewidth}
\centering
\textbf{CIFAR-10}

\begin{tabularx}{\linewidth}{
l
*{4}{>{\centering\arraybackslash}X}
}
\toprule

\rowcolor{HeaderGray}
\textbf{Method}
& \textbf{IF-400}
& \textbf{IF-100}
& \textbf{IF-10}
& \textbf{IF-1} \\

\midrule

Softmax
& 60.32
& 72.20
& 86.72
& 92.98 \\

Softmax + CB
& 61.31
& 73.59
& 86.74
& 92.98 \\

\rowcolor{AoLBlue}
\textbf{Softmax + AoL}
& \textbf{63.94}$^{\mathrm{s}}$
& \textbf{73.74}$^{\mathrm{s}}$
& \textbf{86.80}
& \textbf{93.19} \\

\midrule

Sigmoid
& 60.17
& 71.85
& 86.88
& 93.00 \\

Sigmoid + CB
& 64.45
& \textbf{74.96}
& \textbf{87.28}
& 93.00 \\

\rowcolor{AoLBlue}
\textbf{Sigmoid + AoL}
& \textbf{66.62}
& 72.76
& 87.27
& \textbf{93.39} \\

\bottomrule
\end{tabularx}
\end{minipage}
\end{table}

Overall, AoL generally improves or matches the corresponding baselines across datasets and imbalance levels, with larger gains under severe long-tail settings. Under Softmax, AoL improves CIFAR-100 from $31.50\%$ to $32.43\%$ and CIFAR-10 from $60.32\%$ to $63.94\%$ at IF-$400$, while giving the strongest Softmax results at IF-$100$, IF-$10$, and IF-$1$ on both datasets. Under Sigmoid, AoL likewise improves or matches the baseline across all selected settings, including an increase from $60.17\%$ to $66.62\%$ on CIFAR-10 at IF-$400$. Its effectiveness at IF-$1$ further shows that the resulting training emphasis is not determined by class frequency alone.

\subsection{Online Settings}
\label{sec:exp_online}

We then consider online settings, where data arrive sequentially and the complete dataset is unavailable. We evaluate AoL under two scenarios: streaming learning with relatively stable long-tail distributions and dynamic streaming with evolving long-tail structures. In the first scenario, we construct a long-tailed CIFAR-10 training set and make only $5\%$ of the samples accessible at each step. Following Section~\ref{sec:online_aol}, previously observed samples are maintained in a fixed-capacity buffer using reservoir sampling. Let $n$ be the number of samples observed so far and $K$ the buffer capacity. Once the buffer is full, each new sample is inserted with probability $P_{\mathrm{insert}}=K/n$ and, if inserted, replaces a uniformly sampled buffer element. In the second scenario, a long-tailed subset is dynamically sampled from a balanced source pool according to Equation~(\ref{eq:lt}). A random permutation $\pi_t$ maps long-tail ranks to class identities, so the head-tail assignment evolves over time. Under Period-1, this assignment changes every step, whereas under Period-10 it remains fixed for $10$ consecutive steps before being resampled. We use a learning rate of $0.1$ and Softmax loss reweighting in all online CIFAR-related experiments. At each update, the threshold \(\tau^{(t)}\) is set to the median accuracy of the classes represented in \(\mathcal S^{(t)}\).

The single-seed results for the stable and Period-1 settings are reported in Table~\ref{tab:online_all}, Appendix~\ref{app:offline_online_settings}. Under restricted accessibility, AoL improves CIFAR-10 by $3.32$ and $8.70$ points for IF-$200$ and IF-$100$, respectively, while matching the baseline for IF-$1$. Under Period-1, AoL improves the results for IF-$50$ and IF-$10$ but provides no gain at IF-$200$, indicating that class-level AoL estimates can become unstable when the stream changes too rapidly and observations are sparse. Table~\ref{tab:online_main} reports the three-seed Period-10 evaluation. On CIFAR-10, AoL improves the baseline in all nine seed and IF pairs, with mean gains of $+3.10\pm0.51$, $+1.42\pm1.15$, and $+0.92\pm0.24$ points for IF-$200$, IF-$50$, and IF-$10$, respectively. On CIFAR-100, AoL improves or matches the baseline across all tested seeds, with corresponding mean gains of $+2.31\pm0.89$, $+1.42\pm0.91$, and $+0.02\pm0.03$. These results show that when class-level structure persists over multiple steps, AoL gains are reproducible across random seeds rather than specific to a single run.

\begin{table}[tbp]
\centering
\caption{Period-10 online AoL results on CIFAR-10 and CIFAR-100 over three seeds. Top-1 test accuracy (\%) is reported; the summary rows show mean $\pm$ sample SD.}
\label{tab:online_main}

\scriptsize
\renewcommand{\arraystretch}{1.05}
\setlength{\tabcolsep}{2pt}

\begin{minipage}[t]{0.49\linewidth}
\centering
\textbf{CIFAR-10}

\begin{tabularx}{\linewidth}{
>{\centering\arraybackslash}p{0.08\linewidth}
>{\centering\arraybackslash}p{0.10\linewidth}
>{\centering\arraybackslash}p{0.21\linewidth}
>{\columncolor{AoLBlue}\centering\arraybackslash}p{0.27\linewidth}
>{\centering\arraybackslash}p{0.24\linewidth}
}
\toprule

\rowcolor{HeaderGray}
\textbf{IF}
& \textbf{Seed}
& \textbf{Softmax}
& \cellcolor{HeaderGray}\textbf{Softmax + AoL}
& \textbf{Gain} \\

\midrule

\multirow{4}{*}{200}
& 42
& 77.85
& \textbf{81.44}
& +3.59 \\

& 1024
& 76.52
& \textbf{79.10}
& +2.58 \\

& 2026
& 75.69
& \textbf{78.83}
& +3.14 \\

& Mean
& 76.69$\pm$1.09
& \textbf{79.79$\pm$1.44}
& +3.10$\pm$0.51 \\

\cmidrule(lr){1-5}

\multirow{4}{*}{50}
& 42
& 82.67
& \textbf{85.36}
& +2.69 \\

& 1024
& 84.11
& \textbf{84.55}
& +0.44 \\

& 2026
& 83.29
& \textbf{84.42}
& +1.13 \\

& Mean
& 83.36$\pm$0.72
& \textbf{84.78$\pm$0.51}
& +1.42$\pm$1.15 \\

\cmidrule(lr){1-5}

\multirow{4}{*}{10}
& 42
& 86.93
& \textbf{87.58}
& +0.65 \\

& 1024
& 86.22
& \textbf{87.25}
& +1.03 \\

& 2026
& 86.42
& \textbf{87.51}
& +1.09 \\

& Mean
& 86.52$\pm$0.37
& \textbf{87.45$\pm$0.17}
& +0.92$\pm$0.24 \\

\bottomrule
\end{tabularx}
\end{minipage}
\hfill
\begin{minipage}[t]{0.49\linewidth}
\centering
\textbf{CIFAR-100}

\begin{tabularx}{\linewidth}{
>{\centering\arraybackslash}p{0.08\linewidth}
>{\centering\arraybackslash}p{0.10\linewidth}
>{\centering\arraybackslash}p{0.21\linewidth}
>{\columncolor{AoLBlue}\centering\arraybackslash}p{0.27\linewidth}
>{\centering\arraybackslash}p{0.24\linewidth}
}
\toprule

\rowcolor{HeaderGray}
\textbf{IF}
& \textbf{Seed}
& \textbf{Softmax}
& \cellcolor{HeaderGray}\textbf{Softmax + AoL}
& \textbf{Gain} \\

\midrule

\multirow{4}{*}{200}
& 42
& 42.78
& \textbf{44.18}
& +1.40 \\

& 1024
& 42.61
& \textbf{45.79}
& +3.18 \\

& 2026
& 42.97
& \textbf{45.31}
& +2.34 \\

& Mean
& 42.79$\pm$0.18
& \textbf{45.09$\pm$0.83}
& +2.31$\pm$0.89 \\

\cmidrule(lr){1-5}

\multirow{4}{*}{50}
& 42
& 49.61
& \textbf{50.65}
& +1.04 \\

& 1024
& 48.73
& \textbf{51.19}
& +2.46 \\

& 2026
& 49.95
& \textbf{50.72}
& +0.77 \\

& Mean
& 49.43$\pm$0.63
& \textbf{50.85$\pm$0.29}
& +1.42$\pm$0.91 \\

\cmidrule(lr){1-5}

\multirow{4}{*}{10}
& 42
& \textbf{57.31}
& \textbf{57.31}
& 0.00 \\

& 1024
& \textbf{56.75}
& \textbf{56.75}
& 0.00 \\

& 2026
& 56.63
& \textbf{56.69}
& +0.06 \\

& Mean
& 56.90$\pm$0.36
& \textbf{56.92$\pm$0.34}
& +0.02$\pm$0.03 \\

\bottomrule
\end{tabularx}
\end{minipage}
\end{table}

\paragraph{Large-scale evaluation.}
To further evaluate the performance of AoL on larger-scale datasets, we evaluate Places365~\citep{7968387} and ImageNet-1k~\citep{ILSVRC15} under the same Period-10 protocol. For both datasets, we use the original full training set at $224\times224$ resolution as the source pool, ResNet-34 with seed $42$ on four NVIDIA GH200 Grace Hopper Superchips, a learning rate of $0.01$, and a buffer capacity of $500{,}000$, while all other experimental settings remain unchanged. At IF-$200$, IF-$50$, and IF-$10$, the Softmax baseline/AoL accuracies are $37.23/39.79\%$, $42.31/43.79\%$, and $46.95/47.24\%$ on Places365, yielding gains of $+2.56$, $+1.48$, and $+0.29$ points, respectively. On ImageNet-1k, the corresponding accuracies are $46.74/48.31\%$, $54.46/54.81\%$, and $61.30/61.30\%$, with gains of $+1.57$, $+0.35$, and $0.00$ points. Across both datasets, the gains are larger under more severe imbalance; this is consistent with the trend shown in Table~\ref{tab:online_main}.

\section{Limitations and Future Work}
\label{sec:limitation}

The current AoL formulation intentionally uses binary correctness to isolate error duration and therefore does not capture error severity, confidence, or semantic type. Consequently, persistently mislabeled, ambiguous, or out-of-distribution samples may accumulate large AoL even when emphasizing them does not improve generalization. Class-level aggregation may reduce sensitivity to isolated noise but does not eliminate systematic noise. The online formulation also relies on sufficient temporal consistency: when class-level structure changes too rapidly or observations are sparse, as in the severe Period-1 setting, the accumulated AoL state can become less informative. Our evaluation is intentionally limited to the controlled Softmax/Sigmoid baselines to isolate temporal persistence; otherwise, additional mechanisms and hyperparameters would confound the attribution of AoL. See Appendix~\ref{sec:evaluation_scope} for a fuller discussion of the evaluation scope. Natural extensions include confidence- or uncertainty-aware AoL, robust variants for noisy labels, adaptive temporal aggregation, and task-specific learning-state definitions for structured or generative models.

\section{Conclusion}
\label{sec:conclusion}

We introduced Age of Learning (AoL), a temporal learning-state variable that captures how long prediction errors remain unresolved during training. Unlike instantaneous signals such as loss, margin, and confidence, AoL represents error duration and provides a complementary view of learning difficulty. We developed AoL formulations for both offline and streaming settings, including sample-level and class-level states and lightweight online updates under restricted historical access. Across long-tailed classification experiments, AoL generally improves or matches the corresponding baselines, with reproducible gains in streaming settings where class-level learning difficulty persists over multiple updates. Analysis of class frequency, loss, and margin further shows that AoL is related to conventional difficulty signals but is not determined by them. These results support temporal error persistence as a useful additional signal for characterizing and guiding learning dynamics. Rather than replacing existing measures of difficulty, AoL provides a simple yet effective way to incorporate the history of unresolved prediction errors into adaptive training.



\bibliography{iclr2027_conference}

@InProceedings{Cui_2019_CVPR,
author = {Cui, Yin and Jia, Menglin and Lin, Tsung-Yi and Song, Yang and Belongie, Serge},
title = {Class-Balanced Loss Based on Effective Number of Samples},
booktitle = {Proceedings of the IEEE/CVF Conference on Computer Vision and Pattern Recognition (CVPR)},
month = {June},
year = {2019}
}

@ARTICLE{9380899,
  author={Yates, Roy D. and Sun, Yin and Brown, D. Richard and Kaul, Sanjit K. and Modiano, Eytan and Ulukus, Sennur},
  journal={IEEE Journal on Selected Areas in Communications}, 
  title={Age of Information: An Introduction and Survey}, 
  year={2021},
  volume={39},
  number={5},
  pages={1183-1210},
  doi={10.1109/JSAC.2021.3065072}}

@InProceedings{He_2016_CVPR,
author = {He, Kaiming and Zhang, Xiangyu and Ren, Shaoqing and Sun, Jian},
title = {Deep Residual Learning for Image Recognition},
booktitle = {Proceedings of the IEEE Conference on Computer Vision and Pattern Recognition (CVPR)},
month = {June},
year = {2016}
}

@Techreport{krizhevsky2009learning,
 author = {Krizhevsky, Alex and Hinton, Geoffrey},
 address = {Toronto, Ontario},
 institution = {University of Toronto},
 publisher = {Technical report, University of Toronto},
 title = {Learning multiple layers of features from tiny images},
 year = {2009},
 title_with_no_special_chars = {Learning multiple layers of features from tiny images},
 url = {https://www.cs.toronto.edu/~kriz/learning-features-2009-TR.pdf}
}

@InProceedings{Liu_2019_CVPR,
author = {Liu, Ziwei and Miao, Zhongqi and Zhan, Xiaohang and Wang, Jiayun and Gong, Boqing and Yu, Stella X.},
title = {Large-Scale Long-Tailed Recognition in an Open World},
booktitle = {Proceedings of the IEEE/CVF Conference on Computer Vision and Pattern Recognition (CVPR)},
month = {June},
year = {2019}
}

@inproceedings{NEURIPS2019_621461af,
 author = {Cao, Kaidi and Wei, Colin and Gaidon, Adrien and Arechiga, Nikos and Ma, Tengyu},
 booktitle = {Advances in Neural Information Processing Systems},
 pages = {},
 publisher = {Curran Associates, Inc.},
 title = {Learning Imbalanced Datasets with Label-Distribution-Aware Margin Loss},
 url = {https://proceedings.neurips.cc/paper_files/paper/2019/file/621461af90cadfdaf0e8d4cc25129f91-Paper.pdf},
 volume = {32},
 year = {2019}
}

@inproceedings{NEURIPS2024_d8f5f134,
 author = {Csaba, Botos and Zhang, Wenxuan and M\"{u}ller, Matthias and Lim, Ser-Nam and Elhoseiny, Mohamed and Torr, Philip and Bibi, Adel},
 booktitle = {Advances in Neural Information Processing Systems},
 doi = {10.52202/079017-3813},
 pages = {119976--120012},
 publisher = {Curran Associates, Inc.},
 title = {Label Delay in Online Continual Learning},
 url = {https://proceedings.neurips.cc/paper_files/paper/2024/file/d8f5f134febb4bd74d8f79e338de382c-Paper-Conference.pdf},
 volume = {37},
 year = {2024}
}

@inproceedings{
wang2026streaming,
title={Streaming Stochastic Submodular Maximization with On-Demand User Requests},
author={Honglian Wang and Sijing Tu and Lutz Oettershagen and Aristides Gionis},
booktitle={The Thirty-ninth Annual Conference on Neural Information Processing Systems},
year={2025},
url={https://openreview.net/forum?id=pjfMwXm61w}
}

@inproceedings{NEURIPS2019_e562cd9c,
 author = {Aljundi, Rahaf and Lin, Min and Goujaud, Baptiste and Bengio, Yoshua},
 booktitle = {Advances in Neural Information Processing Systems},
 pages = {},
 publisher = {Curran Associates, Inc.},
 title = {Gradient based sample selection for online continual learning},
 url = {https://proceedings.neurips.cc/paper_files/paper/2019/file/e562cd9c0768d5464b64cf61da7fc6bb-Paper.pdf},
 volume = {32},
 year = {2019}
}

@ARTICLE{10105457,
  author={Zhang, Yifan and Kang, Bingyi and Hooi, Bryan and Yan, Shuicheng and Feng, Jiashi},
  journal={IEEE Transactions on Pattern Analysis and Machine Intelligence}, 
  title={Deep Long-Tailed Learning: A Survey}, 
  year={2023},
  volume={45},
  number={9},
  pages={10795-10816},
  doi={10.1109/TPAMI.2023.3268118}}

@ARTICLE{7364263,
  author={Kam, Clement and Kompella, Sastry and Nguyen, Gam D. and Ephremides, Anthony},
  journal={IEEE Transactions on Information Theory}, 
  title={Effect of Message Transmission Path Diversity on Status Age}, 
  year={2016},
  volume={62},
  number={3},
  pages={1360-1374},
  doi={10.1109/TIT.2015.2511791}}

@INPROCEEDINGS{6195689,
  author={Kaul, Sanjit and Yates, Roy and Gruteser, Marco},
  booktitle={2012 Proceedings IEEE INFOCOM}, 
  title={Real-time status: How often should one update?}, 
  year={2012},
  volume={},
  number={},
  pages={2731-2735},
  doi={10.1109/INFCOM.2012.6195689}}

@Book{bishop2006pattern,
author = {Christopher Bishop},
title = {Pattern Recognition and Machine Learning},
publisher = {Springer},
year = 2006,
pages = {738},
}

@InProceedings{10.1007/978-3-030-65414-6_9,
author="Chou, Hsin-Ping
and Chang, Shih-Chieh
and Pan, Jia-Yu
and Wei, Wei
and Juan, Da-Cheng",
editor="Bartoli, Adrien
and Fusiello, Andrea",
title="Remix: Rebalanced Mixup",
booktitle="Computer Vision -- ECCV 2020 Workshops",
year="2020",
publisher="Springer International Publishing",
address="Cham",
pages="95--110",
isbn="978-3-030-65414-6"
}

@InProceedings{Lin_2017_ICCV,
author = {Lin, Tsung-Yi and Goyal, Priya and Girshick, Ross and He, Kaiming and Dollar, Piotr},
title = {Focal Loss for Dense Object Detection},
booktitle = {Proceedings of the IEEE International Conference on Computer Vision (ICCV)},
month = {Oct},
year = {2017}
}

@InProceedings{pmlr-v267-hacohen25a,
  title = 	 {Predicting the Susceptibility of Examples to Catastrophic Forgetting},
  author =       {Hacohen, Guy and Tuytelaars, Tinne},
  booktitle = 	 {Proceedings of the 42nd International Conference on Machine Learning},
  pages = 	 {21546--21569},
  year = 	 {2025},
  volume = 	 {267},
  series = 	 {Proceedings of Machine Learning Research},
  month = 	 {13--19 Jul},
  publisher =    {PMLR},
  url = 	 {https://proceedings.mlr.press/v267/hacohen25a.html},
}

@ARTICLE{8000687,
  author={Sun, Yin and Uysal-Biyikoglu, Elif and Yates, Roy D. and Koksal, C. Emre and Shroff, Ness B.},
  journal={IEEE Transactions on Information Theory}, 
  title={Update or Wait: How to Keep Your Data Fresh}, 
  year={2017},
  volume={63},
  number={11},
  pages={7492-7508},
  doi={10.1109/TIT.2017.2735804}}

@ARTICLE{8514816,
  author={Kadota, Igor and Sinha, Abhishek and Uysal-Biyikoglu, Elif and Singh, Rahul and Modiano, Eytan},
  journal={IEEE/ACM Transactions on Networking}, 
  title={Scheduling Policies for Minimizing Age of Information in Broadcast Wireless Networks}, 
  year={2018},
  volume={26},
  number={6},
  pages={2637-2650},
  doi={10.1109/TNET.2018.2873606}}

@ARTICLE{8734015,
  author={Kadota, Igor and Sinha, Abhishek and Modiano, Eytan},
  journal={IEEE/ACM Transactions on Networking}, 
  title={Scheduling Algorithms for Optimizing Age of Information in Wireless Networks With Throughput Constraints}, 
  year={2019},
  volume={27},
  number={4},
  pages={1359-1372},
  doi={10.1109/TNET.2019.2918736}}

@inproceedings{
toneva2018an,
title={An Empirical Study of Example Forgetting during Deep Neural Network Learning},
author={Mariya Toneva and Alessandro Sordoni and Remi Tachet des Combes and Adam Trischler and Yoshua Bengio and Geoffrey J. Gordon},
booktitle={International Conference on Learning Representations},
year={2019},
url={https://openreview.net/forum?id=BJlxm30cKm},
}

@inproceedings{
Kang2020Decoupling,
title={Decoupling Representation and Classifier for Long-Tailed Recognition},
author={Bingyi Kang and Saining Xie and Marcus Rohrbach and Zhicheng Yan and Albert Gordo and Jiashi Feng and Yannis Kalantidis},
booktitle={International Conference on Learning Representations},
year={2020},
url={https://openreview.net/forum?id=r1gRTCVFvB}
}

@inproceedings{NEURIPS2023_eeffa70b,
 author = {Shi, Jiang-Xin and Wei, Tong and Xiang, Yuke and Li, Yu-Feng},
 booktitle = {Advances in Neural Information Processing Systems},
 pages = {75669--75687},
 publisher = {Curran Associates, Inc.},
 title = {How Re-sampling Helps for Long-Tail Learning?},
 url = {https://proceedings.neurips.cc/paper_files/paper/2023/file/eeffa70bcbbd43f6bd067edebc6595e8-Paper-Conference.pdf},
 volume = {36},
 year = {2023}
}

@InProceedings{Shrivastava_2016_CVPR,
author = {Shrivastava, Abhinav and Gupta, Abhinav and Girshick, Ross},
title = {Training Region-Based Object Detectors With Online Hard Example Mining},
booktitle = {Proceedings of the IEEE Conference on Computer Vision and Pattern Recognition (CVPR)},
month = {June},
year = {2016}
}

@article{
doi:10.1073/pnas.1611835114,
author = {James Kirkpatrick  and Razvan Pascanu  and Neil Rabinowitz  and Joel Veness  and Guillaume Desjardins  and Andrei A. Rusu  and Kieran Milan  and John Quan  and Tiago Ramalho  and Agnieszka Grabska-Barwinska  and Demis Hassabis  and Claudia Clopath  and Dharshan Kumaran  and Raia Hadsell },
title = {Overcoming catastrophic forgetting in neural networks},
journal = {Proceedings of the National Academy of Sciences},
volume = {114},
number = {13},
pages = {3521-3526},
year = {2017},
doi = {10.1073/pnas.1611835114},
URL = {https://www.pnas.org/doi/abs/10.1073/pnas.1611835114},
eprint = {https://www.pnas.org/doi/pdf/10.1073/pnas.1611835114},
}

@inproceedings{NIPS2017_f8752278,
 author = {Lopez-Paz, David and Ranzato, Marc\textquotesingle Aurelio},
 booktitle = {Advances in Neural Information Processing Systems},
 pages = {},
 publisher = {Curran Associates, Inc.},
 title = {Gradient Episodic Memory for Continual Learning},
 url = {https://proceedings.neurips.cc/paper_files/paper/2017/file/f87522788a2be2d171666752f97ddebb-Paper.pdf},
 volume = {30},
 year = {2017}
}

@inproceedings{NIPS2003_feecee9f,
 author = {Monteleoni, Claire and Jaakkola, Tommi},
 booktitle = {Advances in Neural Information Processing Systems},
 editor = {S. Thrun and L. Saul and B. Sch\"{o}lkopf},
 pages = {},
 publisher = {MIT Press},
 title = {Online Learning of Non-stationary Sequences},
 url = {https://proceedings.neurips.cc/paper_files/paper/2003/file/feecee9f1643651799ede2740927317a-Paper.pdf},
 volume = {16},
 year = {2003}
}

@inproceedings{10.5555/3454287.3455008,
author = {Paszke, Adam and Gross, Sam and Massa, Francisco and Lerer, Adam and Bradbury, James and Chanan, Gregory and Killeen, Trevor and Lin, Zeming and Gimelshein, Natalia and Antiga, Luca and Desmaison, Alban and K\"{o}pf, Andreas and Yang, Edward and DeVito, Zach and Raison, Martin and Tejani, Alykhan and Chilamkurthy, Sasank and Steiner, Benoit and Fang, Lu and Bai, Junjie and Chintala, Soumith},
title = {PyTorch: an imperative style, high-performance deep learning library},
year = {2019},
publisher = {Curran Associates Inc.},
address = {Red Hook, NY, USA},
booktitle = {Proceedings of the 33rd International Conference on Neural Information Processing Systems},
articleno = {721},
numpages = {12}
}

@InProceedings{Zhang_2021_ICCV,
    author    = {Zhang, Xing and Wu, Zuxuan and Weng, Zejia and Fu, Huazhu and Chen, Jingjing and Jiang, Yu-Gang and Davis, Larry S.},
    title     = {VideoLT: Large-Scale Long-Tailed Video Recognition},
    booktitle = {Proceedings of the IEEE/CVF International Conference on Computer Vision (ICCV)},
    month     = {October},
    year      = {2021},
    pages     = {7960-7969}
}

@inproceedings{NEURIPS2024_350e718f,
 author = {Kunstner, Frederik and Milligan, Alan and Yadav, Robin and Schmidt, Mark and Bietti, Alberto},
 booktitle = {Advances in Neural Information Processing Systems},
 doi = {10.52202/079017-0948},
 pages = {30106--30148},
 publisher = {Curran Associates, Inc.},
 title = {Heavy-Tailed Class Imbalance and Why Adam Outperforms Gradient Descent on Language Models},
 url = {https://proceedings.neurips.cc/paper_files/paper/2024/file/350e718ff74062b4bac2c6ffd9e1ac66-Paper-Conference.pdf},
 volume = {37},
 year = {2024}
}

@InProceedings{10.1007/978-3-031-72949-2_18,
author="Sanchez Aimar, Emanuel
and Helgesen, Nathaniel
and Xu, Yonghao
and Kuhlmann, Marco
and Felsberg, Michael",
title="Flexible Distribution Alignment: Towards Long-Tailed Semi-supervised Learning with Proper Calibration",
booktitle="Computer Vision -- ECCV 2024",
year="2025",
publisher="Springer Nature Switzerland",
address="Cham",
pages="307--327",
isbn="978-3-031-72949-2"
}

@InProceedings{Alshammari_2022_CVPR,
    author    = {Alshammari, Shaden and Wang, Yu-Xiong and Ramanan, Deva and Kong, Shu},
    title     = {Long-Tailed Recognition via Weight Balancing},
    booktitle = {Proceedings of the IEEE/CVF Conference on Computer Vision and Pattern Recognition (CVPR)},
    month     = {June},
    year      = {2022},
    pages     = {6897-6907}
}

@inproceedings{
cortes2026improved,
title={Improved Balanced Classification with Theoretically Grounded Loss Functions},
author={Corinna Cortes and Mehryar Mohri and Yutao Zhong},
booktitle={The Thirty-ninth Annual Conference on Neural Information Processing Systems},
year={2025},
url={https://openreview.net/forum?id=SkZUo6Xg61}
}

@InProceedings{pmlr-v267-li25dk,
  title = 	 {Focal-{SAM}: Focal Sharpness-Aware Minimization for Long-Tailed Classification},
  author =       {Li, Sicong and Xu, Qianqian and Yang, Zhiyong and Wang, Zitai and Zhang, Linchao and Cao, Xiaochun and Huang, Qingming},
  booktitle = 	 {Proceedings of the 42nd International Conference on Machine Learning},
  pages = 	 {36624--36651},
  year = 	 {2025},
  volume = 	 {267},
  series = 	 {Proceedings of Machine Learning Research},
  month = 	 {13--19 Jul},
  publisher =    {PMLR},
  url = 	 {https://proceedings.mlr.press/v267/li25dk.html}
}

@InProceedings{pmlr-v139-flaspohler21a,
  title = 	 {Online Learning with Optimism and Delay},
  author =       {Flaspohler, Genevieve E and Orabona, Francesco and Cohen, Judah and Mouatadid, Soukayna and Oprescu, Miruna and Orenstein, Paulo and Mackey, Lester},
  booktitle = 	 {Proceedings of the 38th International Conference on Machine Learning},
  pages = 	 {3363--3373},
  year = 	 {2021},
  volume = 	 {139},
  series = 	 {Proceedings of Machine Learning Research},
  month = 	 {18--24 Jul},
  publisher =    {PMLR},
  url = 	 {https://proceedings.mlr.press/v139/flaspohler21a.html},
}

@InProceedings{10.1007/978-3-030-58604-1_43,
author="Fini, Enrico
and Lathuili{\`e}re, St{\'e}phane
and Sangineto, Enver
and Nabi, Moin
and Ricci, Elisa",
title="Online Continual Learning Under Extreme Memory Constraints",
booktitle="Computer Vision -- ECCV 2020",
year="2020",
publisher="Springer International Publishing",
address="Cham",
pages="720--735",
isbn="978-3-030-58604-1"
}

@inproceedings{
liu2026online,
title={Online Learning and Equilibrium Computation with Ranking Feedback},
author={Mingyang Liu and Yongshan Chen and Zhiyuan Fan and Gabriele Farina and Asuman E. Ozdaglar and Kaiqing Zhang},
booktitle={The Fourteenth International Conference on Learning Representations},
year={2026},
url={https://openreview.net/forum?id=lg6H2oJPky}
}

@inproceedings{
zhang2025nonstationary,
title={Non-stationary Online Learning for Curved Losses: Improved Dynamic Regret via Mixability},
author={Yu-Jie Zhang and Peng Zhao and Masashi Sugiyama},
booktitle={Forty-second International Conference on Machine Learning},
year={2025},
url={https://openreview.net/forum?id=TeHF8YjJaw}
}

@inproceedings{
sridharan2025online,
title={Online Learning with Unknown Constraints},
author={Karthik Sridharan and Seung Won Wilson Yoo},
booktitle={Forty-second International Conference on Machine Learning},
year={2025},
url={https://openreview.net/forum?id=nqQcAhXGSy}
}

@inproceedings{
wu2024stabilizing,
title={Stabilizing Linear Passive-Aggressive Online Learning with Weighted Reservoir Sampling},
author={Skyler Wu and Fred Lu and Edward Raff and James Holt},
booktitle={The Thirty-eighth Annual Conference on Neural Information Processing Systems},
year={2024},
url={https://openreview.net/forum?id=FNOBf6JM7r}
}

@inproceedings{
daniely2026online,
title={Online Learning of Neural Networks},
author={Amit Daniely and Idan Mehalel and Elchanan Mossel},
booktitle={The Thirty-ninth Annual Conference on Neural Information Processing Systems},
year={2025},
url={https://openreview.net/forum?id=oedSIxYWql}
}

@ARTICLE{9137714,
  author={Maatouk, Ali and Kriouile, Saad and Assaad, Mohamad and Ephremides, Anthony},
  journal={IEEE/ACM Transactions on Networking}, 
  title={The Age of Incorrect Information: A New Performance Metric for Status Updates}, 
  year={2020},
  volume={28},
  number={5},
  pages={2215-2228},
  doi={10.1109/TNET.2020.3005549}}

@inproceedings{NEURIPS2022_a39a9ace,
  author    = {Guo, Dandan and Li, Zhuo and Zheng, Meixi and Zhao, He and Zhou, Mingyuan and Zha, Hongyuan},
  title     = {Learning to Re-weight Examples with Optimal Transport for Imbalanced Classification},
  booktitle = {Advances in Neural Information Processing Systems},
  volume    = {35},
  pages     = {25517--25530},
  year      = {2022},
  publisher = {Curran Associates, Inc.},
  doi       = {10.52202/068431-1850},
  url       = {https://proceedings.neurips.cc/paper_files/paper/2022/file/a39a9aceda771cded859ae7560530e09-Paper-Conference.pdf}
}

@inproceedings{pmlr-v80-ren18a,
  author    = {Ren, Mengye and Zeng, Wenyuan and Yang, Bin and Urtasun, Raquel},
  title     = {Learning to Reweight Examples for Robust Deep Learning},
  booktitle = {Proceedings of the 35th International Conference on Machine Learning},
  volume    = {80},
  series    = {Proceedings of Machine Learning Research},
  pages     = {4334--4343},
  year      = {2018},
  publisher = {PMLR},
  url       = {https://proceedings.mlr.press/v80/ren18a.html}
}

@inproceedings{pmlr-v80-katharopoulos18a,
  author    = {Katharopoulos, Angelos and Fleuret, Francois},
  title     = {Not All Samples Are Created Equal: Deep Learning with Importance Sampling},
  booktitle = {Proceedings of the 35th International Conference on Machine Learning},
  volume    = {80},
  series    = {Proceedings of Machine Learning Research},
  pages     = {2525--2534},
  year      = {2018},
  publisher = {PMLR},
  url       = {https://proceedings.mlr.press/v80/katharopoulos18a.html}
}

@inproceedings{yi2021reweighting,
  author    = {Yi, Mingyang and Hou, Lu and Shang, Lifeng and Jiang, Xin and Liu, Qun and Ma, Zhi-Ming},
  title     = {Reweighting Augmented Samples by Minimizing the Maximal Expected Loss},
  booktitle = {International Conference on Learning Representations},
  year      = {2021},
  url       = {https://openreview.net/forum?id=9G5MIc-goqB}
}

@inproceedings{ICLR2025_ded26b34,
  author    = {Sow, Daouda and Woisetschl{\"a}ger, Herbert and Bulusu, Saikiran and Wang, Shiqiang and Jacobsen, Hans Arno and Liang, Yingbin},
  title     = {Dynamic Loss-Based Sample Reweighting for Improved Large Language Model Pretraining},
  booktitle = {International Conference on Learning Representations},
  volume    = {2025},
  pages     = {89490--89520},
  year      = {2025},
  url       = {https://proceedings.iclr.cc/paper_files/paper/2025/file/ded26b348d55953a4863d41540b7d5c4-Paper-Conference.pdf}
}

@inproceedings{NEURIPS2019_e58cc5ca,
  author    = {Shu, Jun and Xie, Qi and Yi, Lixuan and Zhao, Qian and Zhou, Sanping and Xu, Zongben and Meng, Deyu},
  title     = {Meta-Weight-Net: Learning an Explicit Mapping For Sample Weighting},
  booktitle = {Advances in Neural Information Processing Systems},
  volume    = {32},
  year      = {2019},
  publisher = {Curran Associates, Inc.},
  url       = {https://proceedings.neurips.cc/paper_files/paper/2019/file/e58cc5ca94270acaceed13bc82dfedf7-Paper.pdf}
}

@inproceedings{NEURIPS2021_ac56f8fe,
  author    = {Paul, Mansheej and Ganguli, Surya and Dziugaite, Gintare Karolina},
  title     = {Deep Learning on a Data Diet: Finding Important Examples Early in Training},
  booktitle = {Advances in Neural Information Processing Systems},
  volume    = {34},
  pages     = {20596--20607},
  year      = {2021},
  publisher = {Curran Associates, Inc.},
  url       = {https://proceedings.neurips.cc/paper_files/paper/2021/file/ac56f8fe9eea3e4a365f29f0f1957c55-Paper.pdf}
}

@book{Goodfellow-et-al-2016,
    title={Deep Learning},
    author={Ian Goodfellow and Yoshua Bengio and Aaron Courville},
    publisher={MIT Press},
    note={\url{http://www.deeplearningbook.org}},
    year={2016}
}

@article{ILSVRC15,
Author = {Olga Russakovsky and Jia Deng and Hao Su and Jonathan Krause and Sanjeev Satheesh and Sean Ma and Zhiheng Huang and Andrej Karpathy and Aditya Khosla and Michael Bernstein and Alexander C. Berg and Li Fei-Fei},
Title = {{ImageNet Large Scale Visual Recognition Challenge}},
Year = {2015},
journal   = {International Journal of Computer Vision (IJCV)},
doi = {10.1007/s11263-015-0816-y},
volume={115},
number={3},
pages={211-252}
}

@ARTICLE{7968387,
  author={Zhou, Bolei and Lapedriza, Agata and Khosla, Aditya and Oliva, Aude and Torralba, Antonio},
  journal={IEEE Transactions on Pattern Analysis and Machine Intelligence}, 
  title={Places: A 10 Million Image Database for Scene Recognition}, 
  year={2018},
  volume={40},
  number={6},
  pages={1452-1464},
  doi={10.1109/TPAMI.2017.2723009}}

@inproceedings{
zheng2026understanding,
title={Understanding the Dynamics of Forgetting and Generalization in Continual Learning via the Neural Tangent Kernel},
author={Guodong Zheng and Peng Wang and Shengchao Hu and Quan Zheng and Li Shen},
booktitle={The Fourteenth International Conference on Learning Representations},
year={2026},
url={https://openreview.net/forum?id=NE2yIxdo1w}
}

@inproceedings{
jin2026orderdp,
title={Order{DP}: A Theoretically Guaranteed Lossless Dynamic Data Pruning Framework},
author={Chenhan Jin and Shengze Xu and Qingsong Wang and Fan JIA and Dingshuo Chen and Tieyong Zeng},
booktitle={The Fourteenth International Conference on Learning Representations},
year={2026},
url={https://openreview.net/forum?id=e77QyyRQPz}
}
\bibliographystyle{iclr2027_conference}

\clearpage
\appendix

The appendix provides algorithms, additional results, evaluation details, and visualizations of AoL.

\section{Algorithms}
\label{sec:algorithm}

In this section, we provide the algorithmic details of AoL in both offline and online settings. Algorithm~\ref{alg:offline_aol} summarizes the offline procedure described in Section~\ref{sec:approach}. After each training epoch, the model is evaluated on the unaugmented training set, and the sample-level AoL states are updated according to prediction correctness. These states are then aggregated using EMA and converted into normalized sample-level or class-level weights. The resulting weights are used in the following epoch for either loss reweighting or weighted resampling. Algorithm~\ref{alg:online_aol} summarizes the corresponding streaming-setting procedure, where class-level learning states are estimated from the current observation set. For each observed class, its accuracy determines whether the current AoL state increases or resets, while the states of unobserved classes remain unchanged. The resulting class-level AoL states are aggregated using the same EMA principle as in the offline setting, normalized across classes, and used for online loss reweighting. The specific memory policy is independent of the AoL update and determines only the observation set available at each streaming step.

\begin{algorithm}[htbp]
\caption{Offline AoL}
\label{alg:offline_aol}
\begin{algorithmic}[1]
\Require Training set \(\mathcal D=\{(\boldsymbol{x}_i,y_i)\}_{i=1}^{N}\), model \(f_\theta\), number of epochs \(E\), EMA factor \(\rho\), weighting level \(q\in\{\textsc{Sample},\textsc{Class}\}\)
\Ensure AoL-based sample weights \(\{w_i^{(e)}\}_{i=1}^{N}\)
\State Initialize \(a_i^{(0)}\gets1\) and \(w_i^{(0)}\gets1\) for all \(i\in\{1,\dots,N\}\)
\For{\(e=1\) to \(E\)}
    \State Train one epoch using \(\{w_i^{(e-1)}\}_{i=1}^{N}\)
    \State Evaluate \(f_{\theta^{(e)}}\) on the unaugmented training set
    \For{each sample \(i=1,\dots,N\)}
        \State \(\hat y_i^{(e)}\gets\arg\max_{k\in\{1,\dots,C\}}f_{\theta^{(e)}}(\boldsymbol{x}_i)_k\)
        \State \(c_i^{(e)}\gets\mathbb{I}[\hat y_i^{(e)}=y_i]\)
        \If{\(c_i^{(e)}=1\)}
            \State \(a_i^{(e)}\gets1\)
        \Else
            \State \(a_i^{(e)}\gets a_i^{(e-1)}+1\)
        \EndIf
        \If{\(e=1\)}
            \State \(\bar a_i^{(e)}\gets a_i^{(e)}\)
        \Else
            \State \(\bar a_i^{(e)}\gets(1-\rho)\bar a_i^{(e-1)}+\rho a_i^{(e)}\)
        \EndIf
    \EndFor

    \If{\(q=\textsc{Sample}\)}
        \State For each sample \(i=1,\dots,N\), set \(w_i^{(e)}\gets
        \dfrac{\bar a_i^{(e)}}
        {\frac{1}{N}\sum_{j=1}^{N}\bar a_j^{(e)}}\)
    \Else
        \For{each class \(c=1,\dots,C\)}
            \State \(\bar a_c^{(e)}\gets
            \dfrac{1}{|\mathcal I_c|}
            \sum_{i\in\mathcal I_c}\bar a_i^{(e)}\)
        \EndFor
        \State For each class \(c=1,\dots,C\), set \(w_c^{(e)}\gets
        \dfrac{\bar a_c^{(e)}}
        {\frac{1}{C}\sum_{k=1}^{C}\bar a_k^{(e)}}\)
        \State \(w_i^{(e)}\gets w_{y_i}^{(e)}\), for all \(i\)
    \EndIf

    \State Use \(\{w_i^{(e)}\}\) in epoch \(e+1\) for loss reweighting, or set \(p_i^{(e+1)}\propto w_i^{(e)}\) for weighted random sampling
\EndFor
\end{algorithmic}
\end{algorithm}

\begin{algorithm}[htbp]
\caption{Online AoL}
\label{alg:online_aol}
\begin{algorithmic}[1]
\Require Stream \(\{\mathcal B^{(t)}\}_{t=1}^{T}\), model \(f_\theta\), EMA factor \(\rho\), optional memory buffer \(\mathcal M^{(t)}\)
\Ensure Class weights \(\{w_c^{(t)}\}_{c=1}^{C}\)
\State Initialize \(a_c^{(0)}\gets1\) and \(w_c^{(0)}\gets1\) for all \(c\in\{1,\dots,C\}\)
\For{\(t=1\) to \(T\)}
    \State Update the optional buffer \(\mathcal M^{(t)}\) using the chosen memory policy
    \State If a buffer is used, set \(\mathcal S^{(t)}\gets\mathcal M^{(t)}\); otherwise, set \(\mathcal S^{(t)}\gets\mathcal B^{(t)}\)
    \State Evaluate the current model on \(\mathcal S^{(t)}\) and compute \(r_c^{(t)}\) according to Eq.~(\ref{eq:online_class_accuracy})
    \State Set \(\tau^{(t)}\gets\operatorname{median}\{r_c^{(t)}:c\text{ is represented in }\mathcal S^{(t)}\}\)
    \For{each class \(c\) represented in \(\mathcal S^{(t)}\)}
        \If{\(r_c^{(t)}<\tau^{(t)}\)}
            \State \(a_c^{(t)}\gets a_c^{(t-1)}+1\)
        \Else
            \State \(a_c^{(t)}\gets1\)
        \EndIf
    \EndFor

    \State Keep \(a_c^{(t)}\gets a_c^{(t-1)}\) for classes not represented in \(\mathcal S^{(t)}\)
    \State Apply the EMA aggregation in Eqs.~(\ref{eq:ema_init}--\ref{eq:ema_aol}) to the class-level states \(\{a_c^{(t)}\}\), yielding \(\{\bar a_c^{(t)}\}\)
    \State For each class \(c=1,\dots,C\), set \(w_c^{(t)}\gets
    \dfrac{\bar a_c^{(t)}}
    {\frac{1}{C}\sum_{k=1}^{C}\bar a_k^{(t)}}\)
    \State Assign each observed sample the weight of its ground-truth class for online loss reweighting
\EndFor
\end{algorithmic}
\end{algorithm}

\section{Additional Experimental Details}
\label{app:additional_results}

\subsection{Offline and Online Settings}
\label{app:offline_online_settings}

In this subsection, we provide additional offline and online results. Table~\ref{tab:offline_all} presents additional offline results corresponding to Table~\ref{tab:offline_main}. Table~\ref{tab:offline_all} reports additional offline results on CIFAR-100 and CIFAR-10 under different imbalance factors (IFs), including both loss reweighting and weighted resampling. For each setting, we evaluate the corresponding CB baselines together with the sample-level and class-level AoL variants and report the better AoL result, using the class-level result when their accuracies are equal. Table~\ref{tab:online_all} reports additional single-seed results under both the access-constrained and evolving long-tail streaming settings. These CIFAR-10 experiments use ResNet-32, class-level AoL, a fixed learning rate of \(0.1\), and the same buffer and optimization protocols described in Section~\ref{sec:exp_online}. In the access-constrained setting, the stream exposes only 5\% of the long-tailed training samples at each epoch, while in the dynamic setting the long-tail structure is resampled over time from a balanced IF-1 source pool. These results further show that AoL benefits from temporally persistent class-level difficulty and sufficiently reliable buffer statistics.

\begin{table}[htbp]
\centering
\caption{Offline results on CIFAR-100 and CIFAR-10 with varying imbalance factors (IFs). Top-1 test accuracy (\%) is reported. Panel (a) evaluates loss reweighting, and panel (b) evaluates weighted resampling. The better result between the sample- and class-level AoL is presented in the table; the class-level result is reported when they are the same. Superscript $\mathrm{s}$ denotes that the result is achieved at the sample-level; the results without the superscript are obtained at the class-level. In addition, \(\beta\) is the CB hyperparameter and \(\rho\) is the AoL factor. CB$^\dagger$ denotes the CB variant with \(\beta=0.0\), and AoL$^\dagger$ denotes the AoL variant with \(\rho=0.0\).}
\label{tab:offline_all}

\scriptsize
\setlength{\tabcolsep}{2.5pt}
\renewcommand{\arraystretch}{0.99}

\newcommand{\param}[1]{\textcolor{black!55}{#1}}

\begin{tabularx}{\linewidth}{
>{\raggedright\arraybackslash}p{0.18\linewidth}
*{8}{>{\centering\arraybackslash}X}
}

\multicolumn{9}{c}{\textbf{(a) Loss reweighting}} \\
\addlinespace[0.25em]
\toprule

\multicolumn{9}{c}{\textbf{CIFAR-100}} \\
\midrule

\rowcolor{HeaderGray}
\textbf{Method}
& \textbf{IF-400}
& \textbf{IF-300}
& \textbf{IF-200}
& \textbf{IF-100}
& \textbf{IF-50}
& \textbf{IF-20}
& \textbf{IF-10}
& \textbf{IF-1} \\
\midrule

Softmax
& 31.50 & \textbf{33.46} & 35.77 & 40.10
& 44.85 & 51.26 & 57.15 & 71.59 \\

Softmax + CB
& 32.04 & 32.54 & 35.83 & 40.40
& \textbf{45.38} & \textbf{52.38} & 57.48 & 71.59 \\

\multicolumn{1}{c}{\param{\(\beta\)}}
& \param{0.9}
& \param{0.9}
& \param{0.9}
& \param{0.9}
& \param{0.9}
& \param{0.99}
& \param{0.99}
& \param{--} \\

\rowcolor{AoLBlue}
\textbf{Softmax + AoL}
& \textbf{32.43}
& \textbf{33.46}
& \textbf{36.08}
& \textbf{40.51}
& 44.90
& 51.87
& \textbf{57.98}
& \textbf{71.67} \\

\multicolumn{1}{c}{\param{\(\rho\)}}
& \param{0.3}
& \param{0.0}
& \param{0.1}
& \param{0.8}
& \param{0.5}
& \param{0.3}
& \param{0.9}
& \param{0.2} \\

\midrule

Sigmoid
& \textbf{33.00}
& 34.34
& 36.21
& 40.65
& 45.33
& 52.21
& 57.63
& 71.33 \\

Sigmoid + CB
& 32.86
& 34.13
& 36.67
& 40.81
& 45.03
& \textbf{52.98}
& \textbf{58.44}
& 71.33 \\

\multicolumn{1}{c}{\param{\(\beta\)}}
& \param{0.9}
& \param{0.9}
& \param{0.9}
& \param{0.9}
& \param{0.9}
& \param{0.99}
& \param{0.99}
& \param{--} \\

\rowcolor{AoLBlue}
\textbf{Sigmoid + AoL}
& \textbf{33.00}
& \textbf{34.50}$^{\mathrm{s}}$
& \textbf{36.70}$^{\mathrm{s}}$
& \textbf{40.84}
& \textbf{45.41}
& 52.42
& 58.07
& \textbf{71.56} \\

\multicolumn{1}{c}{\param{\(\rho\)}}
& \param{0.0}
& \param{\(0.3^{\mathrm{s}}\)}
& \param{\(0.1^{\mathrm{s}}\)}
& \param{0.1}
& \param{1.0}
& \param{1.0}
& \param{0.7}
& \param{0.2} \\

\addlinespace[0.20em]
\midrule
\multicolumn{9}{c}{\textbf{CIFAR-10}} \\
\midrule

\rowcolor{HeaderGray}
\textbf{Method}
& \textbf{IF-400}
& \textbf{IF-300}
& \textbf{IF-200}
& \textbf{IF-100}
& \textbf{IF-50}
& \textbf{IF-20}
& \textbf{IF-10}
& \textbf{IF-1} \\
\midrule

Softmax
& 60.32
& 65.70
& 67.94
& 72.20
& 77.03
& 82.78
& 86.72
& 92.98 \\

Softmax + CB
& 61.31
& 65.21
& \textbf{69.56}
& 73.59
& \textbf{78.14}
& \textbf{84.18}
& 86.74
& 92.98 \\

\multicolumn{1}{c}{\param{\(\beta\)}}
& \param{0.99}
& \param{0.9}
& \param{0.99}
& \param{0.9999}
& \param{0.999}
& \param{0.999}
& \param{0.99}
& \param{--} \\

\rowcolor{AoLBlue}
\textbf{Softmax + AoL}
& \textbf{63.94}$^{\mathrm{s}}$
& \textbf{66.80}
& 67.94
& \textbf{73.74}$^{\mathrm{s}}$
& 77.89
& 83.40
& \textbf{86.80}
& \textbf{93.19} \\

\multicolumn{1}{c}{\param{\(\rho\)}}
& \param{\(0.6^{\mathrm{s}}\)}
& \param{0.2}
& \param{0.0}
& \param{\(0.8^{\mathrm{s}}\)}
& \param{0.5}
& \param{0.4}
& \param{1.0}
& \param{0.6} \\

\midrule

Sigmoid
& 60.17
& 62.84
& 65.76
& 71.85
& 77.40
& \textbf{83.92}
& 86.88
& 93.00 \\

Sigmoid + CB
& 64.45
& \textbf{67.67}
& \textbf{70.42}
& \textbf{74.96}
& \textbf{79.69}
& \textbf{83.92}
& \textbf{87.28}
& 93.00 \\

\multicolumn{1}{c}{\param{\(\beta\)}}
& \param{0.999}
& \param{0.99}
& \param{0.9}
& \param{0.9999}
& \param{0.9999}
& \param{0.9}
& \param{0.9999}
& \param{--} \\

\rowcolor{AoLBlue}
\textbf{Sigmoid + AoL}
& \textbf{66.62}
& 67.46
& 68.17
& 72.76
& 78.50
& \textbf{83.92}
& 87.27
& \textbf{93.39} \\

\multicolumn{1}{c}{\param{\(\rho\)}}
& \param{0.2}
& \param{1.0}
& \param{0.2}
& \param{0.2}
& \param{0.9}
& \param{0.0}
& \param{0.8}
& \param{0.8} \\

\addlinespace[0.45em]
\midrule
\multicolumn{9}{c}{\textbf{(b) Weighted resampling}} \\
\addlinespace[0.20em]
\midrule

\multicolumn{9}{c}{\textbf{CIFAR-100}} \\
\midrule

\rowcolor{HeaderGray}
\textbf{Method}
& \textbf{IF-400}
& \textbf{IF-300}
& \textbf{IF-200}
& \textbf{IF-100}
& \textbf{IF-50}
& \textbf{IF-20}
& \textbf{IF-10}
& \textbf{IF-1} \\
\midrule

Softmax
& 31.50
& \textbf{33.46}
& 35.77
& 40.10
& 44.85
& 51.26
& 57.15
& \textbf{71.59} \\

Softmax + CB$^\dagger$
& 31.78
& 33.18
& \textbf{36.21}
& \textbf{40.69}
& \textbf{45.22}
& 51.80
& 57.28
& 70.70 \\

Softmax + CB
& 31.50
& 33.13
& 35.33
& 40.27
& 44.17
& 51.89
& 57.17
& 70.70 \\

\multicolumn{1}{c}{\param{\(\beta\)}}
& \param{0.9}
& \param{0.9}
& \param{0.9}
& \param{0.9}
& \param{0.9}
& \param{0.9}
& \param{0.9}
& \param{--} \\

Softmax + AoL$^\dagger$
& 31.78
& 33.18
& \textbf{36.21}
& \textbf{40.69}
& \textbf{45.22}
& 51.80
& 57.28
& 70.70 \\

\rowcolor{AoLBlue}
\textbf{Softmax + AoL}
& \textbf{32.26}
& 33.18
& \textbf{36.21}
& \textbf{40.69}
& \textbf{45.22}
& \textbf{52.13}
& \textbf{57.55}
& 70.84 \\

\multicolumn{1}{c}{\param{\(\rho\)}}
& \param{0.2}
& \param{0.0}
& \param{0.0}
& \param{0.0}
& \param{0.0}
& \param{0.4}
& \param{0.5}
& \param{0.6} \\

\midrule

Sigmoid
& \textbf{33.00}
& \textbf{34.34}
& 36.21
& 40.65
& \textbf{45.33}
& 52.21
& 57.63
& \textbf{71.33} \\

Sigmoid + CB$^\dagger$
& 32.65
& 33.54
& 36.56
& 40.55
& 44.91
& 52.20
& 57.91
& 71.26 \\

Sigmoid + CB
& 31.90
& 34.04
& \textbf{36.84}
& 40.33
& 44.48
& 52.03
& \textbf{58.08}
& 71.26 \\

\multicolumn{1}{c}{\param{\(\beta\)}}
& \param{0.9}
& \param{0.9}
& \param{0.9}
& \param{0.9}
& \param{0.9}
& \param{0.9}
& \param{0.99}
& \param{--} \\

Sigmoid + AoL$^\dagger$
& 32.65
& 33.54
& 36.56
& 40.55
& 44.91
& 52.20
& 57.91
& 71.26 \\

\rowcolor{AoLBlue}
\textbf{Sigmoid + AoL}
& 32.65
& 33.96
& 36.56
& \textbf{40.98}
& 45.32
& \textbf{52.53}
& 58.02
& 71.26 \\

\multicolumn{1}{c}{\param{\(\rho\)}}
& \param{0.0}
& \param{0.2}
& \param{0.0}
& \param{0.6}
& \param{0.8}
& \param{0.9}
& \param{0.4}
& \param{0.0} \\

\addlinespace[0.20em]
\midrule

\multicolumn{9}{c}{\textbf{CIFAR-10}} \\
\midrule

\rowcolor{HeaderGray}
\textbf{Method}
& \textbf{IF-400}
& \textbf{IF-300}
& \textbf{IF-200}
& \textbf{IF-100}
& \textbf{IF-50}
& \textbf{IF-20}
& \textbf{IF-10}
& \textbf{IF-1} \\
\midrule

Softmax
& 60.32
& 65.70
& 67.94
& 72.20
& 77.03
& 82.78
& 86.72
& 92.98 \\

Softmax + CB$^\dagger$
& 59.33
& 63.23
& \textbf{69.20}
& 70.52
& 77.53
& 82.55
& 86.75
& 92.91 \\

Softmax + CB
& \textbf{63.17}
& \textbf{68.25}
& 67.17
& \textbf{73.13}
& 77.91
& 83.28
& 86.75
& 92.91 \\

\multicolumn{1}{c}{\param{\(\beta\)}}
& \param{0.99}
& \param{0.99}
& \param{0.999}
& \param{0.99}
& \param{0.99}
& \param{0.9999}
& \param{0.9}
& \param{--} \\

Softmax + AoL$^\dagger$
& 59.33
& 63.23
& \textbf{69.20}
& 70.52
& 77.53
& 82.55
& 86.75
& 92.91 \\

\rowcolor{AoLBlue}
\textbf{Softmax + AoL}
& 62.96
& 65.55$^{\mathrm{s}}$
& \textbf{69.20}
& 72.87$^{\mathrm{s}}$
& \textbf{78.80}
& \textbf{83.75}
& \textbf{86.77}
& \textbf{93.16} \\

\multicolumn{1}{c}{\param{\(\rho\)}}
& \param{0.7}
& \param{\(0.4^{\mathrm{s}}\)}
& \param{0.0}
& \param{\(0.3^{\mathrm{s}}\)}
& \param{0.9}
& \param{0.9}
& \param{0.5}
& \param{0.5} \\

\midrule

Sigmoid
& 60.17
& 62.84
& 65.76
& 71.85
& 77.40
& \textbf{83.92}
& 86.88
& 93.00 \\

Sigmoid + CB$^\dagger$
& 59.25
& 61.46
& 67.16
& 71.89
& \textbf{78.47}
& 83.22
& 86.95
& 92.82 \\

Sigmoid + CB
& 62.97
& 63.89
& 68.98
& \textbf{73.30}
& 77.97
& 83.90
& 86.95
& 92.82 \\

\multicolumn{1}{c}{\param{\(\beta\)}}
& \param{0.999}
& \param{0.99}
& \param{0.999}
& \param{0.99}
& \param{0.9}
& \param{0.99}
& \param{0.9, 0.9999}
& \param{--} \\

Sigmoid + AoL$^\dagger$
& 59.25
& 61.46
& 67.16
& 71.89
& \textbf{78.47}
& 83.22
& 86.95
& 92.82 \\

\rowcolor{AoLBlue}
\textbf{Sigmoid + AoL}
& \textbf{63.87}
& \textbf{65.94}
& \textbf{69.38}$^{\mathrm{s}}$
& 72.58
& \textbf{78.47}
& 83.90
& \textbf{87.14}
& \textbf{93.26}$^{\mathrm{s}}$ \\

\multicolumn{1}{c}{\param{\(\rho\)}}
& \param{0.5}
& \param{0.5}
& \param{\(0.7^{\mathrm{s}}\)}
& \param{0.7}
& \param{0.0}
& \param{0.3}
& \param{0.9}
& \param{\(0.5^{\mathrm{s}}\)} \\

\bottomrule
\end{tabularx}

\end{table}

\begin{table}[htbp]
\centering
\caption{Online AoL results on CIFAR-10. Top-1 test accuracy (\%) is reported. Panel (a) evaluates relatively stable long-tail streams where 5\% of the training samples are randomly selected at each epoch; the large-buffer setting retains all observed samples, while the limited-buffer setting uses reservoir sampling with \(K=10000\). Panel (b) evaluates evolving long-tail streams generated from a balanced IF\(=1\) source pool with a limited buffer of \(K=50000\). AoL generally improves or matches Softmax, with larger gains under stronger imbalance and limited-buffer access. Period-10 also shows more consistent gains than Period-1.}

\label{tab:online_all}

\scriptsize
\setlength{\tabcolsep}{2.5pt}
\renewcommand{\arraystretch}{1.05}

\newcommand{\onlineparam}[1]{\textcolor{black!55}{#1}}

\begin{tabularx}{\linewidth}{
>{\raggedright\arraybackslash}p{0.18\linewidth}
*{8}{>{\centering\arraybackslash}X}
}

\multicolumn{9}{c}{\textbf{(a) Online streams with 5\% random selection at each epoch}} \\
\addlinespace[0.25em]
\toprule

&
\multicolumn{4}{c}{\textbf{Large buffer (all observed)}}
&
\multicolumn{4}{c}{\textbf{Limited buffer \(K=10000\)}} \\
\cmidrule(lr){2-5}
\cmidrule(lr){6-9}

\rowcolor{HeaderGray}
\textbf{Method}
& \textbf{IF-200}
& \textbf{IF-100}
& \textbf{IF-50}
& \textbf{IF-1}
& \textbf{IF-200}
& \textbf{IF-100}
& \textbf{IF-50}
& \textbf{IF-1} \\
\midrule

Softmax
& 30.31
& 35.00
& 40.80
& \textbf{80.76}
& 30.31
& 35.00
& 40.80
& \textbf{80.76} \\

\rowcolor{AoLBlue}
\textbf{Softmax + AoL}
& \textbf{34.15}
& \textbf{39.53}
& \textbf{44.66}
& \textbf{80.76}
& \textbf{33.63}
& \textbf{43.70}
& \textbf{45.19}
& \textbf{80.76} \\

\multicolumn{1}{c}{\onlineparam{\(\rho\)}}
& \onlineparam{0.2}
& \onlineparam{0.1}
& \onlineparam{0.1, 0.2}
& \onlineparam{0.0}
& \onlineparam{0.2}
& \onlineparam{0.1}
& \onlineparam{0.1}
& \onlineparam{0.0} \\

\bottomrule
\end{tabularx}

\begin{tabularx}{\linewidth}{
>{\raggedright\arraybackslash}p{0.18\linewidth}
*{10}{>{\centering\arraybackslash}X}
}

\multicolumn{11}{c}{\textbf{(b) Online dynamic long-tail streams with limited buffer size \(K=50000\)}} \\
\addlinespace[0.25em]
\toprule

&
\multicolumn{5}{c}{\textbf{Dynamic Long-Tail with Period-1}}
&
\multicolumn{5}{c}{\textbf{Dynamic Long-Tail with Period-10}} \\
\cmidrule(lr){2-6}
\cmidrule(lr){7-11}

\rowcolor{HeaderGray}
\textbf{Method}
& \textbf{IF-200}
& \textbf{IF-100}
& \textbf{IF-50}
& \textbf{IF-20}
& \textbf{IF-10}
& \textbf{IF-200}
& \textbf{IF-100}
& \textbf{IF-50}
& \textbf{IF-20}
& \textbf{IF-10} \\
\midrule

Softmax
& \textbf{54.03}
& 75.66
& 81.58
& 86.38
& 87.49
& 77.85
& 79.77
& 82.67
& 86.41
& 86.93 \\

\rowcolor{AoLBlue}
\textbf{Softmax + AoL}
& \textbf{54.03}
& \textbf{76.45}
& \textbf{84.46}
& \textbf{86.40}
& \textbf{87.60}
& \textbf{81.44}
& \textbf{82.49}
& \textbf{85.36}
& \textbf{86.42}
& \textbf{87.58} \\

\multicolumn{1}{c}{\onlineparam{\(\rho\)}}
& \onlineparam{0.0}
& \onlineparam{0.1}
& \onlineparam{0.2}
& \onlineparam{0.4}
& \onlineparam{0.6}
& \onlineparam{0.1}
& \onlineparam{0.5}
& \onlineparam{0.4}
& \onlineparam{0.9}
& \onlineparam{0.8} \\

\bottomrule
\end{tabularx}
\end{table}

\subsection{Arrival Imbalance}
\label{app:arrival_imbalance}

In this subsection, we study a complementary offline setting that isolates arrival imbalance from class-frequency imbalance. Standard epoch-wise training with shuffle already induces a mild form of natural arrival sparsity, since samples from all classes are not guaranteed to appear uniformly in every batch. To amplify this effect, we introduce an arrival loader that, at the beginning of each epoch, uniformly retains a fraction \(p \in \{1,2,5,10,20,50,100\}\%\) of the training samples. This construction induces a more temporally sparse condition: although the dataset remains class-balanced, some samples may not appear for several consecutive epochs. The retained subset of each epoch is the only data used for optimization and AoL updates in that epoch. We evaluate this setting on balanced CIFAR-100 and balanced CIFAR-10 using loss reweighting only, with the same ResNet-32 architecture, SGD optimizer, batch size, learning-rate schedule, momentum, and weight decay as in the previous offline experiments. For this arrival-specific extension of AoL, the age update is computed only from the samples that actually enter the current epoch training subset; samples that were previously observed but do not appear in the current epoch are incremented by one, while samples never observed remain NaN and are excluded from the aggregation.

Table~\ref{tab:arrival_imbalance}(a) and Table~\ref{tab:arrival_imbalance}(b) report balanced CIFAR-100 and CIFAR-10 results, respectively. The main purpose of this experiment is to examine whether AoL remains informative when class frequencies are unchanged but sample arrival becomes increasingly sparse over time. The results suggest that AoL is generally more helpful in the low-arrival regime, especially under the 1\%--5\% settings where supervision is temporally sparse. This may indicate that AoL captures not only class frequency but also the duration for which samples remain absent from optimization. As the arrival rate increases, the performance gap between AoL and the baseline generally becomes smaller, with several settings showing comparable performance. This suggests that the benefit of an explicit AoL signal may be most visible when temporal persistence of errors is a key limiting factor. Overall, the experiment suggests that AoL captures a complementary aspect of difficulty beyond static class balance, namely the persistence of under-exposure during training.

\begin{table}[tbp]
\centering
\caption{Offline arrival-imbalance ablation on balanced CIFAR-100 and CIFAR-10. Top-1 test accuracy (\%) is reported. At the beginning of each epoch, an arrival loader uniformly retains a fraction \(p \in \{1,2,5,10,20,50,100\}\%\) of the training set, while all other optimization settings remain identical to the standard offline setup. AoL generally improves or matches both Softmax and Sigmoid baselines across arrival fractions, with the largest gains under sparse arrivals.}
\label{tab:arrival_imbalance}

\scriptsize
\setlength{\tabcolsep}{2.5pt}
\renewcommand{\arraystretch}{1.05}

\newcommand{\arrivalparam}[1]{\textcolor{black!55}{#1}}

\begin{tabularx}{\linewidth}{
>{\raggedright\arraybackslash}p{0.20\linewidth}
*{7}{>{\centering\arraybackslash}X}
}

\multicolumn{8}{c}{\textbf{(a) CIFAR-100 (Balanced)}} \\
\addlinespace[0.25em]
\toprule

\rowcolor{HeaderGray}
\textbf{Method}
& \textbf{1\%}
& \textbf{2\%}
& \textbf{5\%}
& \textbf{10\%}
& \textbf{20\%}
& \textbf{50\%}
& \textbf{100\%} \\
\midrule

Softmax
& 18.55
& 32.04
& 53.24
& \textbf{58.67}
& \textbf{66.74}
& 69.88
& 71.59 \\

\rowcolor{AoLBlue}
\textbf{Softmax + AoL}
& \textbf{19.43}
& \textbf{34.27}
& \textbf{53.66}
& \textbf{58.67}
& \textbf{66.74}
& \textbf{69.92}
& \textbf{71.67} \\

\multicolumn{1}{c}{\arrivalparam{\(\rho\)}}
& \arrivalparam{0.4}
& \arrivalparam{0.4}
& \arrivalparam{0.7}
& \arrivalparam{0.0}
& \arrivalparam{0.0}
& \arrivalparam{0.6}
& \arrivalparam{0.2} \\

\midrule

Sigmoid
& 19.05
& \textbf{35.55}
& 54.35
& 59.31
& 66.19
& 69.76
& 71.33 \\

\rowcolor{AoLBlue}
\textbf{Sigmoid + AoL}
& \textbf{20.01}
& \textbf{35.55}
& \textbf{54.49}
& \textbf{59.73}
& \textbf{66.66}
& \textbf{70.04}
& \textbf{71.56} \\

\multicolumn{1}{c}{\arrivalparam{\(\rho\)}}
& \arrivalparam{0.4}
& \arrivalparam{0.0}
& \arrivalparam{1.0}
& \arrivalparam{0.9}
& \arrivalparam{0.7}
& \arrivalparam{0.2}
& \arrivalparam{0.2} \\

\bottomrule

\addlinespace[0.55em]

\multicolumn{8}{c}{\textbf{(b) CIFAR-10 (Balanced)}} \\
\addlinespace[0.25em]
\toprule

\rowcolor{HeaderGray}
\textbf{Method}
& \textbf{1\%}
& \textbf{2\%}
& \textbf{5\%}
& \textbf{10\%}
& \textbf{20\%}
& \textbf{50\%}
& \textbf{100\%} \\
\midrule

Softmax
& 57.04
& \textbf{75.11}
& 84.97
& 86.69
& 90.46
& 91.88
& 92.98 \\

\rowcolor{AoLBlue}
\textbf{Softmax + AoL}
& \textbf{61.54}
& \textbf{75.11}
& \textbf{85.66}
& \textbf{87.49}
& \textbf{90.60}
& \textbf{92.62}
& \textbf{93.19} \\

\multicolumn{1}{c}{\arrivalparam{\(\rho\)}}
& \arrivalparam{0.9}
& \arrivalparam{0.0}
& \arrivalparam{0.2}
& \arrivalparam{0.1}
& \arrivalparam{0.3}
& \arrivalparam{0.7}
& \arrivalparam{0.6} \\

\midrule

Sigmoid
& 62.55
& 76.58
& 84.79
& 86.82
& 90.02
& \textbf{92.42}
& 93.00 \\

\rowcolor{AoLBlue}
\textbf{Sigmoid + AoL}
& \textbf{63.80}
& \textbf{78.12}
& \textbf{85.64}
& \textbf{87.56}
& \textbf{90.59}
& \textbf{92.42}
& \textbf{93.39} \\

\multicolumn{1}{c}{\arrivalparam{\(\rho\)}}
& \arrivalparam{0.4}
& \arrivalparam{1.0}
& \arrivalparam{0.5}
& \arrivalparam{1.0}
& \arrivalparam{0.6}
& \arrivalparam{0.0}
& \arrivalparam{0.8} \\

\bottomrule
\end{tabularx}

\end{table}

\section{Evaluation Scope and Experimental Design}
\label{sec:evaluation_scope}

\paragraph{Controlled baseline design.}
The primary goal of the experiments is to determine whether temporal error persistence provides a useful learning signal, rather than to rank AoL as a complete long-tailed recognition system. AoL is deliberately implemented as a lightweight training signal that can be attached to standard objectives. We therefore use standard Softmax and Sigmoid losses as unweighted references and CB Loss~\citep{Cui_2019_CVPR} as the principal frequency-based weighting reference. This comparison is particularly informative because CB and AoL modify the same part of the training pipeline, the emphasis assigned to training examples, while deriving that emphasis from fundamentally different information. Here, CB uses static class statistics through the effective number of samples, whereas AoL uses the temporal persistence of prediction errors. The backbone, base loss family, data construction, and optimization pipeline can therefore remain coherent, allowing the contribution of the learning signal itself to be examined with limited changes.

This design should not be interpreted as an attempt to establish a leaderboard ranking against state-of-the-art long-tailed recognition systems. Contemporary approaches may jointly modify representation learning, classifier training, logits or margins, sampling strategies, auxiliary objectives, or multi-stage optimization. \emph{Comparing such complete systems would conflate the proposed temporal signal with several additional design choices and would answer a different question from the one studied here.} These methods are included in the related-work discussion to position AoL within the broader literature, whereas the empirical comparisons are intentionally structured to isolate the new learning signal. Similarly, loss reweighting and weighted resampling are used as controlled mechanisms for applying AoL rather than being presented as the methodological novelty themselves.

\paragraph{Hyperparameter selection and reporting.}
We follow the experimental convention of the original CB Loss study~\citep{Cui_2019_CVPR}, in which hyperparameters are selected by cross-validation. For CB, we use the original candidate set
\(\beta\in\{0.9,0.99,0.999,0.9999\}\).
For AoL, we use the predefined grid
\(\rho\in\{0.0,0.1,\ldots,1.0\}\)
under the same validation protocol. Sample-level and class-level AoL are treated as two predefined formulations rather than post-hoc modifications; the formulation selected by validation is reported, with the class-level formulation used in case of a tie. Test data are reserved for final accuracy reporting. To make the selection transparent, Table~\ref{tab:offline_all} reports the corresponding \(\beta\) and \(\rho\) values, and the superscript $\mathrm{s}$ explicitly identifies cases in which the sample-level AoL formulation performs better.

\paragraph{Experimental coverage.}
Although the principal comparison is intentionally controlled, the evaluation varies the factors that are directly relevant to the proposed temporal signal. The offline study covers CIFAR-10 and CIFAR-100, eight imbalance factors, Softmax and Sigmoid objectives, loss reweighting and weighted resampling, and both sample-level and class-level AoL formulations. The streaming study considers access-constrained streams, large and limited memory buffers, rapidly evolving Period-1 streams, more temporally persistent Period-10 streams, CIFAR-10 and CIFAR-100, and three random seeds in the main Period-10 evaluation. We further evaluate the same Period-10 formulation on the full Places365 and ImageNet-1k source pools and separately study balanced-data arrival sparsity and diagnostic AoL trajectories. These experiments are intended to test the proposed signal across imbalance severity, temporal persistence, observation constraints, dataset scale, and intervention mechanism rather than across unrelated system components.

\paragraph{Streaming scope and comparison protocol.}
The streaming experiments consider task-free classification over a fixed label space whose sample accessibility and class-frequency structure evolve over time. They are therefore distinct from task-incremental or class-incremental continual-learning protocols that introduce explicit task boundaries, new classes, delayed supervision, or additional task-specific assumptions. In the controlled streaming comparisons, AoL and the Softmax baseline use the same incoming data, memory policy, buffer capacity, architecture, optimizer, and training schedule; the difference is whether the class-level temporal state is used for loss reweighting. This matched design isolates the effect of AoL under the streaming conditions studied here. Specialized continual-learning systems remain compatible directions for incorporating AoL, but comparing complete systems with different replay, regularization, and task assumptions is outside the controlled question addressed in the present experiments.

\paragraph{Online computation and memory.}
An online AoL update is performed once after each training epoch in all reported experiments, rather than after every SGD mini-batch. At an AoL update, the current model is evaluated on the observation set \(\mathcal{S}^{(t)}\) to estimate class-level accuracy and update the AoL states. This requires a forward-only evaluation and no additional backward pass, while the persistent AoL state itself requires only \(O(C)\) memory beyond the data buffer. The computational cost therefore depends on both the observation set size and the chosen AoL evaluation frequency. The 500,000-sample buffer used in the large-scale Places365 and ImageNet-1k experiments is intended to evaluate scalability under a common large-scale Period-10 protocol, rather than to minimize memory demand. Memory-constrained behavior is examined separately through the limited-buffer CIFAR experiments, including \(K=10{,}000\) and \(K=50{,}000\).

\paragraph{Interpretation of ``complementary.''}
We use the term \emph{complementary} to describe the information represented by AoL: class frequency describes the amount of available data, and loss or margin describes the current prediction state, whereas AoL describes how long an incorrect state has persisted. The correlation analysis in Section~\ref{sec:exp_observation} directly examines this distinction. Complementarity in this sense does not imply that AoL must produce an additive gain when mechanically combined with every existing rebalancing method. Combining AoL with CB, for example, would require an additional design choice for composing temporal and frequency-based weights, together with corresponding normalization and hyperparameter selection. The main experiments therefore evaluate the signals separately so that the contribution of temporal persistence can be identified without introducing an additional combination rule.

\paragraph{Statistical scope of the results.}
The extensive offline tables are intended primarily as controlled mechanism studies across datasets, imbalance factors, base losses, weighting levels, and intervention mechanisms. Accordingly, small differences in single-run offline results are not interpreted as evidence of statistical significance. We use terms such as ``matches'' or ``comparable'' when the observed difference is negligible. Reproducibility across random initialization is examined explicitly in the three-seed Period-10 experiments in Table~\ref{tab:online_main}, where all individual seed results, means, and sample standard deviations are reported. This separates broad controlled coverage from the dedicated multi-seed robustness evaluation.

\paragraph{Large-scale evaluation and dataset scope.}
The Places365 and ImageNet-1k experiments use the original full training sets at \(224\times224\) resolution as source pools and apply the same controlled Period-10 construction used in the streaming study. Their purpose is to test whether the AoL mechanism remains effective when the source dataset and memory buffer are substantially larger, without dataset-specific changes to the method. These experiments therefore provide evidence of computational and dataset-scale transfer under the studied controlled imbalance protocol; they are not presented as a state-of-the-art comparison on naturally occurring long-tailed benchmarks. Natural long-tail distributions constitute an additional evaluation axis that is orthogonal to the present controlled study.

\paragraph{Diagnostic visualizations.}
The heatmaps in Section~\ref{sec:visualization} are intentionally generated from standard Softmax training in which AoL is logged but not fed back into optimization. This isolates the learning-state signal itself: an AoL-guided model would change the trajectory being diagnosed and would therefore mix the existence of the temporal pattern with the effect of the intervention. The CIFAR-10 visualization uses a power-law color normalization with \(\gamma=0.35\), \(v_{\min}=1\), and \(v_{\max}=100\), while the CIFAR-100 visualization uses \(\gamma=0.5\) with the same bounds. Raw AoL values are not numerically clipped before visualization; values above \(100\) simply saturate at the upper end of the displayed color scale. The nonlinear normalization is used only to preserve visual contrast among low- and medium-AoL regions and has no effect on any quantitative result.

\paragraph{Reproducibility.}
The footnote link on the first page of the main text contains the code used for the reported offline and streaming AoL updates, long-tail and streaming data construction, hyperparameter configurations, and memory-buffer handling. Together with the reported architectures, optimization schedules, random seeds, selected \(\beta\) and \(\rho\) values, buffer capacities, and data-construction rules, this is intended to make the controlled comparisons independently reproducible.

\section{Diagnostic Visualizations}
\label{sec:visualization}

To better understand the training dynamics captured by AoL, we visualize the aggregated class-level AoL trajectories on CIFAR-10 and CIFAR-100 under the long-tailed setting with imbalance factor $200$. These visualizations are purely diagnostic: AoL is computed and logged from the training process, but it is not fed back into the optimization objective, nor is it used as a weighting term or sampling probability during training. We use a nonlinear color normalization in the heatmaps so that both low- and medium-AoL regions remain visually distinguishable.

\begin{figure}[htbp]
    \centering
    \includegraphics[width=\linewidth]{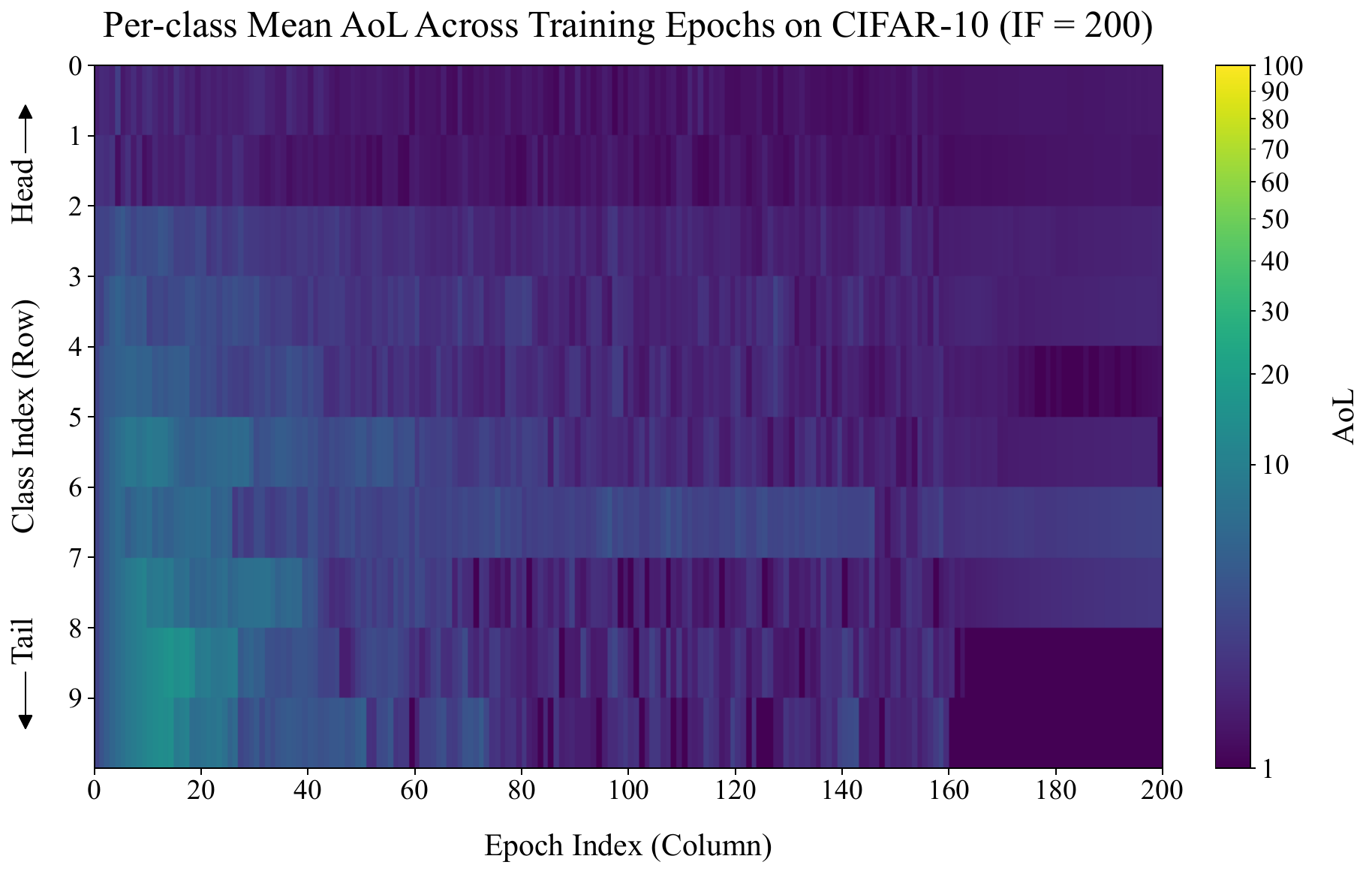}
    \caption{Per-class mean AoL across training epochs on CIFAR-10 (IF = 200) in the Softmax setting. The head classes are shown at the top and the tail classes at the bottom. AoL gradually decreases over training, and the later stage becomes noticeably darker, indicating reduced prediction-error persistence as optimization proceeds.}
    \label{fig:aol_cifar10_heatmap}
\end{figure}

Figure~\ref{fig:aol_cifar10_heatmap} shows the CIFAR-10 result. The head classes remain relatively dark and stable throughout training, whereas the tail classes exhibit noticeably larger AoL values in the early stage and more visible fluctuations over time. Although AoL is not used to guide optimization in this diagnostic experiment, the overall AoL level decreases as training progresses, suggesting that persistent error episodes become shorter over training. In the later stage of training, especially after approximately 160 epochs, the heatmap becomes visibly darker, and the separation between earlier and later training phases becomes more pronounced. This transition is consistent with the shift to a lower learning rate, although the exact boundary should be interpreted as a qualitative observation rather than a strict phase transition.

\begin{figure}[H]
    \centering
    \includegraphics[width=\linewidth]{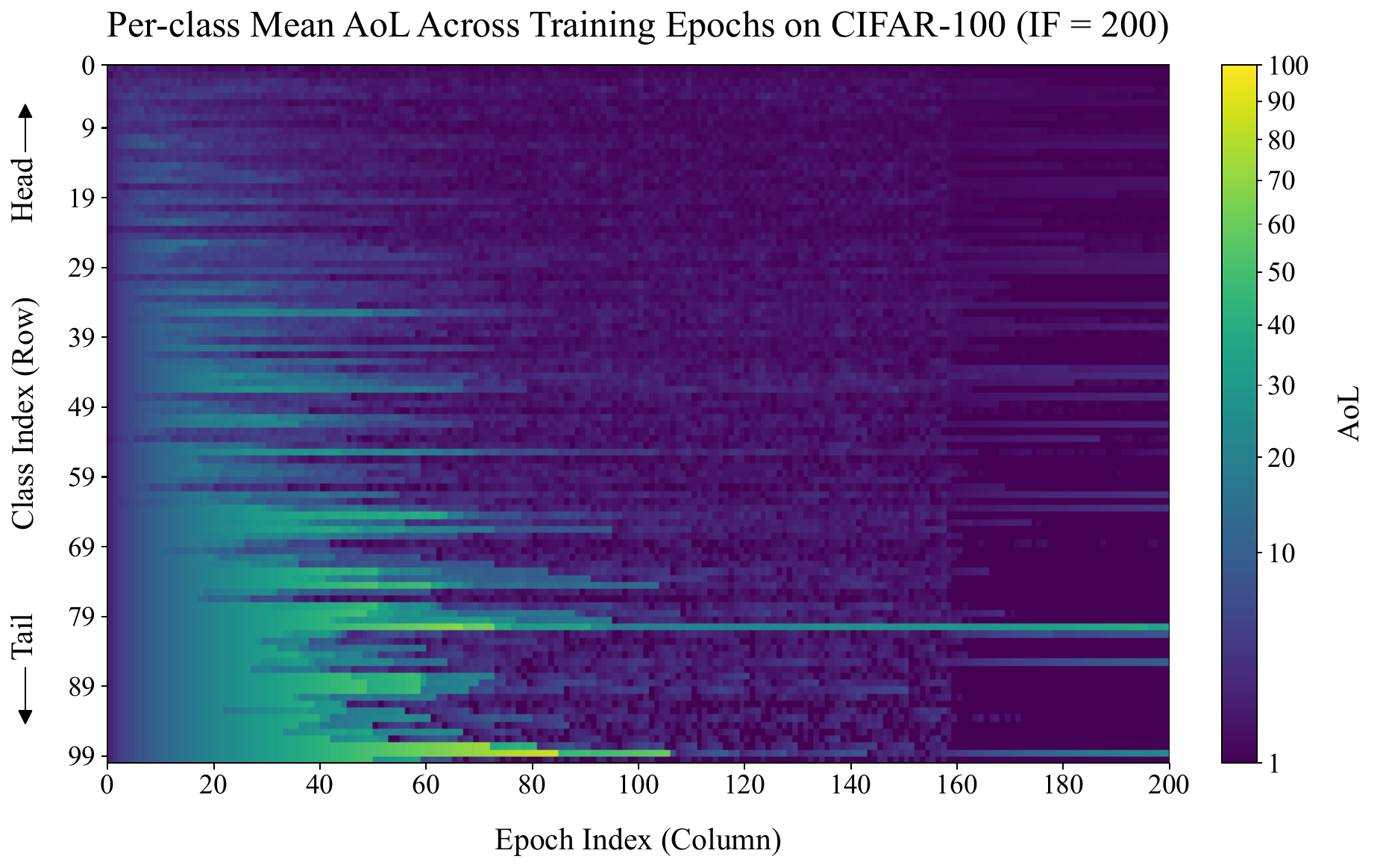}
    \caption{Per-class mean AoL across training epochs on CIFAR-100 (IF = 200) in the Softmax setting. Compared with CIFAR-10, the head-tail gap is more pronounced, and tail classes exhibit more persistent high-AoL regions, especially in the early and middle stages of training. The late-stage darkening is consistent with the transition to a lower learning rate.}
    \label{fig:aol_cifar100_heatmap}
\end{figure}

Figure~\ref{fig:aol_cifar100_heatmap} presents the corresponding CIFAR-100 visualization. Compared with CIFAR-10, CIFAR-100 exhibits a more pronounced head and tail difference: head classes generally stabilize earlier, while many tail classes retain higher AoL for longer. Several tail classes still show intermittent bright bands in the middle and late stages, indicating that their predictions remain less stable than those of head classes. Nevertheless, the global trend is still downward, and the latter part of training is substantially darker than the early part. This suggests that AoL captures a meaningful notion of learning persistence even when it is not explicitly used for optimization. That is, AoL can reveal class-dependent learning dynamics that are not fully reflected by instantaneous accuracy alone.

\end{document}